\documentclass[sn-basic,iicol]{sn-jnl}
\usepackage{graphicx}
\usepackage{multirow}
\usepackage{amsmath,amssymb,amsfonts}
\usepackage{amsthm}
\usepackage{mathrsfs}
\usepackage[title]{appendix}
\usepackage{color}
\usepackage[table,dvipsnames]{xcolor}
\usepackage{textcomp}
\usepackage{manyfoot}
\usepackage{booktabs}
\usepackage{pifont}
\usepackage[linesnumbered,ruled,vlined]{algorithm2e}
\usepackage{algpseudocode}
\SetKwInput{KwInput}{Input} 
\newcommand{\xmark}{\ding{55}}
\usepackage{bm}
\usepackage{tikz}
\usepackage{tabularx}
\usepackage{colortbl}
\usepackage{stmaryrd}
\usepackage{flushend}
\usepackage{makecell}
\usepackage{xcolor}
\usepackage{tabstackengine}
\usepackage{twemojis}
\usepackage[accsupp]{axessibility}
\usepackage{subcaption}

\usetikzlibrary{arrows.meta}

\usepackage[T1]{fontenc}
\usepackage{soul}

\usepackage[capitalize]{cleveref}

\usepackage{xspace}

\makeatletter
\DeclareRobustCommand\onedot{\futurelet\@let@token\@onedot}
\def\@onedot{\ifx\@let@token.\else.\null\fi\xspace}

\def\eg{\textit{e.g}\onedot} 
\def\ie{\textit{i.e}\onedot} 
\def\cf{\textit{cf}\onedot} 
\def\etc{\textit{etc}\onedot} \def\vs{\textit{vs}\onedot}

\makeatother

\theoremstyle{thmstyleone}

\theoremstyle{thmstyletwo}

\theoremstyle{thmstylethree}

\newcommand{\EncI}{E_\text{img}}
\newcommand{\EncT}{E_\text{txt}}

\newcommand{\mb}[1]{\ensuremath{\mathbf{#1}}}
\newcommand{\mc}[1]{\ensuremath{\mathcal{#1}}}
\newcommand{\bs}[1]{\ensuremath{\boldsymbol{#1}}}

\newcommand{\src}[1]{{#1}_\text{s}}
\newcommand{\trg}[1]{{#1}_\text{t}}
\newcommand{\stot}[1]{{#1}_{\text{s}\shortrightarrow\text{t}}}

\newcommand{\emb}[1]{\bar{#1}}

\newcommand{\vartn}[1]{{\footnotesize\,$\pm$#1}}
\newcommand{\query}[1]{\texttt{#1}}

\newcommand{\prompt}{\mathsf{P}}
\newcommand{\promptConcept}{\mathsf{P}^{*}}
\newcommand{\targetFeat}{\trg{\emb{\mb{f}}}}
\newcommand{\freeze}{\texttwemoji{2744}}
\newcommand{\concept}{\mathsf{S}^{*}}

\definecolor{srccolor}{RGB}{0,102,0}
\definecolor{trgcolor}{RGB}{99,0,99}
\definecolor{intermcolor}{RGB}{153,0,153}
\definecolor{darkred}{RGB}{153,0,0}
\definecolor{darkblue}{RGB}{0, 0, 153}

\definecolor{nightcolor}{rgb}{0.0, 0.0, 0.8}
\definecolor{snowcolor}{rgb}{0.2, 0.6, 1.0}
\definecolor{gamecolor}{rgb}{1.0, 0.48, 0.48}
\definecolor{grayprompt}{rgb}{0.82, 0.82, 0.82}
\definecolor{colorcell}{rgb}{0.819, 0.94, 0.956}

\definecolor{deeppurple}{RGB}{99,0,99}

\definecolor{colrelprompt}{rgb}{0.0, 0.4, 0.6}
\definecolor{colirrprompt}{rgb}{0.8, 0.0, 0.0}

\newcommand{\relprompt}{{\color{colrelprompt!100}relevant prompt}}
\newcommand{\irrprompt}{{\color{colirrprompt!100}irrelevant prompt}}

\definecolor{codeblue}{rgb}{0.25,0.5,0.5}

\SetCommentSty{mycommfont}

\newcommand{\method}{P{\O}DA\xspace}
\newcommand{\methodone}{PIDA\xspace}
\newcommand{\methodconcept}{P{\O}DA-concept\xspace}

\newcommand{\DAsetting}[2]{{#1}$\shortrightarrow${#2}}

\newcommand{\dashrule}[1][black]{
  \color{#1}\rule[\dimexpr.5ex-.2pt]{4pt}{.4pt}\xleaders\hbox{\rule{4pt}{0pt}\rule[\dimexpr.5ex-.2pt]{4pt}{.4pt}}\hfill\kern0pt
}
\newcommand{\rulecolor}[1]{
  \def\CT@arc@{\color{#1}}
}

\usetikzlibrary{arrows.meta}

\begin{document}

\title[Article Title]{\centering
{Domain Adaptation with a Single Vision-Language Embedding}
}

\author*[1]{\fnm{Mohammad} \sur{Fahes}}\email{mohammad.fahes@inria.fr}

\author[1,2]{\fnm{Tuan-Hung} \sur{Vu}}\email{tuan-hung.vu@valeo.com}

\author[1,2]{\fnm{Andrei} \sur{Bursuc}}\email{andrei.bursuc@valeo.com}

\author[3]{\fnm{Patrick} \sur{Pérez}}\email{patrick@kyutai.org}

\author[1]{\fnm{Raoul} \sur{de Charette}}\email{raoul.de-charette@inria.fr}

\affil[1]{\orgdiv{Inria}, 
\city{Paris},
\country{France}}

\affil[2]{\orgdiv{Valeo.ai}, 
\city{Paris}, 
\country{France}}

\affil[3]{\orgdiv{Kyutai}, 
\city{Paris},  
\country{France}}

\abstract{Domain adaptation has been extensively investigated in computer vision but still requires access to target data at the training time, which might be difficult to obtain in real-world autonomous driving scenarios, especially under rare or adverse conditions. In this paper, we present a new framework for domain adaptation relying on a single Vision-Language~(VL) latent embedding instead of full target data. First, leveraging a contrastive language-image pre-training model (CLIP), we propose prompt/photo-driven instance normalization (PIN). PIN is a feature augmentation method that mines multiple visual styles using a single target VL latent embedding, by optimizing affine transformations of low-level source features. The VL embedding can come from a language prompt describing the target domain, a partially optimized language prompt, or a single unlabeled target image. Second, we show that these mined styles (\ie, augmentations) can be used for zero-shot (\ie, target-free) and one-shot unsupervised domain adaptation. Experiments on semantic segmentation in real-world driving datasets, including Cityscapes and ACDC (adverse conditions), demonstrate the effectiveness of the proposed method, which outperforms relevant baselines in the practical zero-shot and one-shot settings.}

\keywords{Zero-shot domain adaptation, vision-language models, feature augmentation, style mining, prompt/photo-driven instance normalization}

\maketitle

\section{Introduction}
\label{sec:intro}

The success of deep learning in image classification~\citep{krizhevsky2012imagenet} has led to a swift paradigm shift over the past decade. This shift has propelled the rapid evolution of algorithms, with deep neural networks (DNNs) forming the backbone of nearly every modern computer vision method. For instance, supervised semantic segmentation methods have achieved remarkable success in improving high-resolution predictions~\citep{long2015fully,chen2017deeplab,chen2018encoder,cheng2020panoptic, wang2020deep}, incorporating multi-scale processing~\citep{zhao2017pyramid,lin2017feature} and enhancing computational efficiency~\citep{zhao2018icnet}. Yet, DNN-based methods are still far from being reliably applied in many scenarios for critical applications like real-world autonomous driving. 
The reason is that learning-based systems suppose the training and testing data to be independent and identically distributed (i.i.d.), a hypothesis that is violated most of the times as the training data tend to under-represent the true generative distribution. Consequently, in controlled settings where segmentation models are trained using data from the targeted operational design domains, the accuracy can meet the high industry-level expectations on in-domain data; yet, when tested on out-of-distribution data, these models often undergo drastic performance drops~\citep{ovadia2019can}.

To mitigate the so-called ``domain-shift'' problem~\citep{ben2010theory}, unsupervised domain adaptation~(UDA)~\citep{ganin2016domain,sun2016deep,hoffman2018cycada,zou2018unsupervised,tsai2018learning,vu2019advent} has been proposed; and aims at training on labeled data from a \textit{source} domain and unlabeled data from the \textit{target} domain. This alleviates the need for annotating data, which is often laborious, taking 2 to 3 hours per image on average \citep{cordts2016cityscapes,sakaridis2021acdc}. 
Moreover, modern architectures can require massive amounts of annotated images~\citep{kirillov2023segment}.
UDA is thus seen as a label-efficient framework for dealing with domain shifts.

Despite its apparent simplicity, even gathering unlabeled data can be challenging under certain conditions. For example, as driving through fire or sandstorm rarely occurs in real life and can be dangerous, capturing these conditions is not trivial.
One may argue on using internet images for UDA.
However, in the industrial context, the practice of using public data is limited or forbidden.
Recent works aim to reduce the burden of target data collection campaigns by devising one-shot unsupervised domain adaptation (OSUDA)~\citep{luo2020adversarial,wu2022style} methods, \ie, using one target image for training, which substitutes label-efficiency by a more challenging data-efficiency setting.
\begin{figure}[t]
	\centering
	\includegraphics[width=1.0\linewidth]{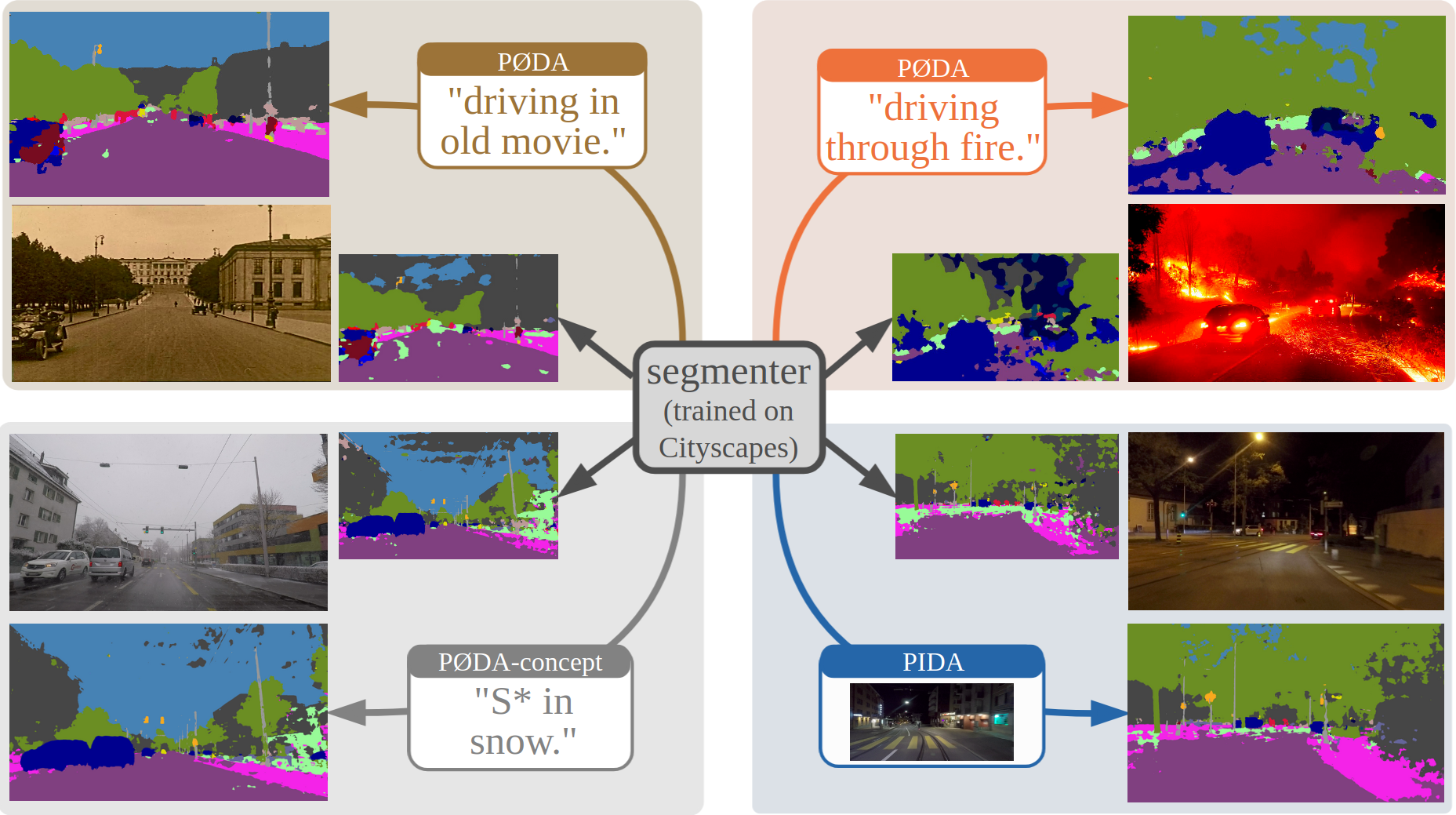}
	\caption{\small\textbf{Domain adaptation 
    using a single vision-language (VL) embedding.}
	The proposed framework adapts a segmenter model (here DeepLabv3+), initially trained on the source domain (Cityscapes), to unseen target conditions using only one embedding in a shared VL space. (\textit{Top}) \method{} performs adaptation via a single text prompt. (\textit{Bottom-left}) \methodconcept{} employs a prompt where the concept $\concept{}$, optimized from source images, is combined with textual descriptions of target conditions. (\textit{Bottom-right}) \methodone{} adapts the model using a single unlabeled target image. Smaller segmentation masks adjacent to each test image show predictions from the source-only model for comparison.
    }
	\label{fig:teaser}
\end{figure}

In this paper, we frame the challenging new task of domain adaptation using a single Vision-Language (VL) latent embedding. At training time, our method adapts the segmentation model to the domain defined by either a single description in natural language (\ie, a \textit{prompt}), a partially optimized prompt, or a single unlabeled target image.
This makes the method suitable for zero-shot domain adaptation when the embedding comes from a prompt, and for one-shot unsupervised domain adaptation (OSUDA) when it comes from an unlabeled target image. Since our approach leverages a single VL target embedding which may originate from a prompt or an image, we coin it \textit{Prompt-driven Zero-shot Domain Adaptation}~(\method{}) when using a fully verbalized target prompt,~\textit{\methodconcept{}} when a part of the target prompt is optimized using source images, and \textit{Photo-driven one-shot Domain Adaptation} (\methodone{}\footnote{In \methodone{} the ``I'' stands for the roman number 1.}) when using a target image.

\cref{fig:teaser} outlines the primary goal of our work with a few qualitative examples.
Without seeing any fire or old movie images during training, and using a simple description of these conditions with a language prompt (\ie, \method), the adapted models succeed in segmenting out critical scene objects, exhibiting fewer errors than the original source-only model. 
\cref{fig:teaser} additionally shows the improvements brought by our framework when part of the prompt is optimized from source data (\ie, \methodconcept) or when the VL embedding comes from an unlabeled target image (\ie, \methodone). 

\begin{figure*}[t!]
    \centering
    \includegraphics[width=1.0\linewidth]{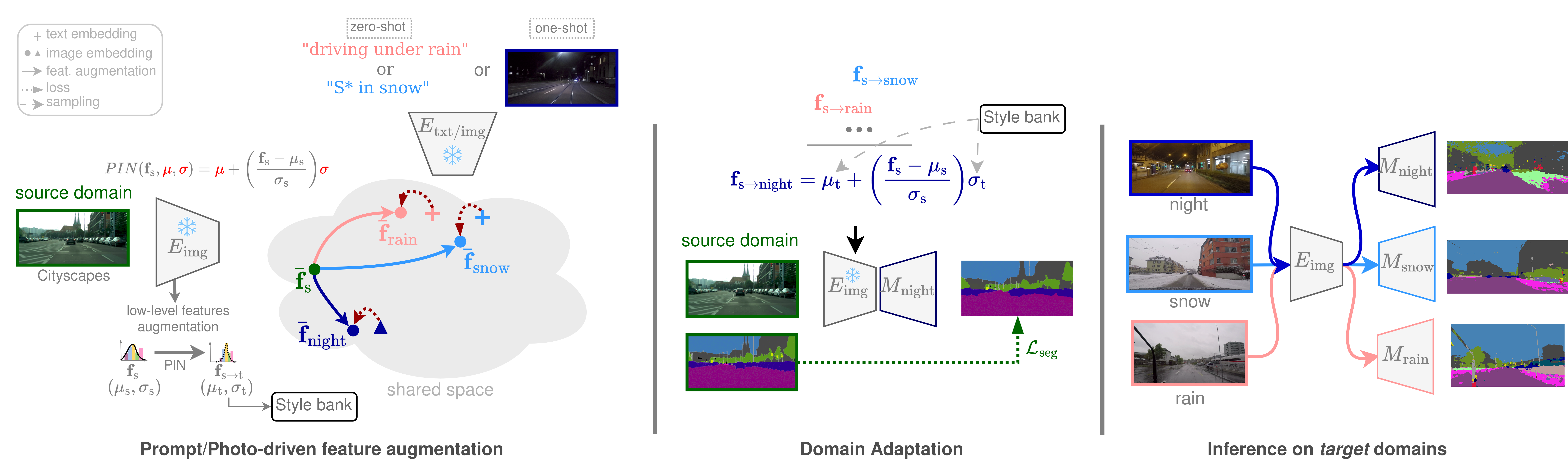}
    \vspace{-0.2cm}
    \caption{
 \small \textbf{Overview of our framework for domain adaptation using a single vision-language (VL) embedding.} (\textit{Left})~Given a single VL embedding representing the target domain \textemdash{} either from a text prompt {\large \textbf{+}} or a target image {\large $\blacktriangle$} \textemdash{} we use a frozen ResNet encoder initialized with CLIP weights to optimize low-level feature transformations from source to target. The transformation is based on our proposed prompt/photo- driven instance normalization (PIN), which optimizes the two vectors $\mu$ and $\sigma$ shown in red. These are initialized from the source domain ($\mu_s$ and $\sigma_s$) and optimized such that the latent features better align with the target VL embedding ({\large \textbf{+}} or {\large $\blacktriangle$}). The resulting parameters ($\mu_t$,$\sigma_t$) are saved in a style bank. A detailed description of PIN is provided in~\cref{fig:method}. (\textit{Middle})~Zero-shot or one-shot unsupervised domain adaptation is then achieved by fine-tuning the segmentation model ($M$) on features that are affinely transformed using styles sampled from the corresponding style bank (\eg, $\color{darkblue}\textbf{f}_{\text{s}\shortrightarrow\text{night}}$). (\textit{Right})~This enables inference on previously unseen target domains.
 }

    \vspace{-0.2cm}
\label{fig:overview}
\end{figure*}

Our method, illustrated in \cref{fig:overview},  
is made possible by leveraging the VL connections from the seminal CLIP model \citep{radford2021learning}. 
Trained on 400M web-crawled image-text pairs, CLIP has revolutionized multi-modal representation learning, bringing outstanding transfer capability to tasks such as image synthesis~\citep{kwon2022clipstyler,gal2021stylegan,patashnik2021styleclip}, multi-modal fusion~\citep{jatavallabhula2023conceptfusion}, semantic segmentation~\citep{li2022language, zhou2022extract}, few-shot learning~\citep{zhou2022learning,gao2024clip,fahes2026clip}, as well as open-vocabulary object detection~\citep{minderer2022simple}. 

Our work exploits CLIP latent space and proposes a simple and effective method that converts source-domain embeddings into target-domain ones (\cref{fig:overview}, left), by optimizing style-specific components of low-level features. This procedure can be seen as a specific form of feature \textit{augmentation} utilizing a single VL embedding guidance coming from either a prompt or an image. 
Fine-tuning the segmentation model using the optimized styles (\cref{fig:overview}, middle) helps mitigating the distribution gap between the two domains,
thus improving the performance on target domains (\cref{fig:overview}, right).

This paper extends our previous work~\citep{fahes2023poda}, \method{}, and proposes a more general formulation. The extension is threefold. 
	\textit{(i)} We further analyze the effectiveness of relevant prompts in \method{} showing that it originates from the combined description of content (\eg, ``driving'') and style (\eg, ``at night''). 
	\textit{(ii)} Inspired by prompt learning~\citep{zhou2022learning,zhou2022conditional}, we show 
 that rather than encompassing content and style in the prompt, the content \textit{concept} $\concept{}$ can be 
 optimized from source data and combined with a textual style (\eg, ``$\concept{}$ at night''). 
	\textit{(iii)}~
 Since our framework requires only a single VL embedding, we propose \methodone{}, which utilizes an unlabeled image instead of a prompt, thus allowing adaptation to conditions that are hard to describe with a textual prompt.
	These novel contributions not only improve the original \method{} but also extend its scope of application.

Overall, our contributions are the following: 
	\begin{itemize}
		\item We define a novel task of domain adaptation, which aims at adapting a source-trained model to a target domain provided \textit{only} a single VL embedding, and demonstrate its effectiveness with either a prompt or an image.
		\item Unlike other CLIP-based methods that navigate CLIP latent space using direct image representations, we alter only the features, without relying on the appearance in the pixel space. We argue that this is particularly useful for downstream tasks such as semantic segmentation, where good features are decisive for the performance of the segmentation head. Specifically, we introduce prompt/photo-driven instance normalization (PIN), which optimizes affine transformations of the low-level source features such that their embeddings match that of a single image/text condition describing the unseen target domain.
		\item We propose Prompt-driven Zero-shot Domain Adaptation~\citep{fahes2023poda} (\method), which utilizes a \textit{single prompt} to describe the content and style of the target condition, demonstrating its performance on a wide variety of scenarios: (i) from clear weather/daytime to adverse conditions (snow, rain, night), (ii) from synthetic to real, (iii) from real to synthetic.
		\item We introduce the novel \methodconcept, building on a \textit{concept optimization} strategy to relax the necessity of describing the content of the source images using language. Experiments show that when the optimized concept is coupled with a visual appearance (\ie, style) description, the downstream performance is significantly improved.
		\item We introduce \methodone which instead utilizes the VL embedding of a \textit{single image} to adapt a trained model. For peculiar visual conditions which are difficult to describe with text prompts, we demonstrate that it further improves the performance of our method.
	\end{itemize}

\section{Related works}

\smallskip\noindent\textbf{Unsupervised Domain Adaptation (UDA).~}
{The UDA literature is vast and encompasses different yet connected approaches: adversarial learning~\citep{ganin2016domain,tsai2018learning}, self-training~\citep{zou2019confidence,li2019bidirectional}, entropy minimization~\citep{vu2019advent,pan2020unsupervised}, generative-based adaptation~\citep{hoffman2018cycada}, etc. The domain gap is commonly reduced at the level of the input~\citep{hoffman2018cycada,yang2020fda}, of the features~\citep{ganin2016domain,sun2016deep,wang2017deep,long2018conditional} or of the output~\citep{tsai2018learning,vu2019advent,pan2020unsupervised}.}

{Recently, the more challenging setting of One-Shot Unsupervised Domain Adaptation (OSUDA) has been proposed.  
~\citet{luo2020adversarial} show that traditional UDA methods fail when only a single unlabeled target image is available.
To mitigate the risk of overfitting on the style of the single available image, 
the authors propose a style mining algorithm, based on both a stylized image generator and a task-specific module.
~\citet{wu2022style} introduce an approach based on style mixing and patch-wise prototypical matching (SM-PPM).
During training, channel-wise mean and standard deviation of a randomly sampled source image's features  are linearly mixed with the target ones.
Patch-wise prototypical matching helps addressing negative adaptation~\citep{li2020content}.}

{In the more challenging zero-shot setting (where no target image is available), ~\citet{lengyel2021zero} tackle day-to-night domain adaptation using physics priors. They introduce a color invariant convolution layer (CIConv) that is added to make the network invariant to different lighting conditions. We note that this zero-shot adaptation method
is orthogonal to ours and is restricted to a specific type of domain gap.}

\smallskip\noindent\textbf{Text-driven image synthesis.~}
{Recently, contrastive vision-language pretraining has shown unprecedented success for multimodal learning in several downstream tasks such as zero-shot classification~\citep{radford2021learning}, multi-modal retrieval~\citep{jia2021scaling} and visual question answering~\citep{li2021align}.
This encouraged the community to edit images using text descriptions, a task that was previously challenging due to the gap between vision and language representations.
For example, StyleCLIP~\citep{patashnik2021styleclip} uses prompts to optimize StyleGAN~\citep{karras2019style} latent vectors and guide the generation process.
However, the generation is limited to the training distribution of StyleGAN.
To overcome this issue, StyleGAN-NADA~\citep{gal2021stylegan} utilizes CLIP embeddings of text prompts to perform domain adaptation of the generator, which is in this case trainable.} 

{For text-guided style transfer, CLIPstyler~\citep{kwon2022clipstyler} does not rely on a generative process. This setting is more realistic for not being restricted to a specific distribution, and challenging at the same time for the use of the encapsulated information in CLIP latent space. Indeed, there is no one-to-one mapping between image and text representations and regularization is needed to extract the useful information from a text embedding. Thus, in the same work~\citep{kwon2022clipstyler}, a U-net~\citep{ronneberger2015u} autoencoder that preserves the content is optimized while the output image embedding in CLIP latent space is varying during the optimization process.}

{We note that a common point in prior works is the mapping from pixel-space to CLIP latent space during the optimization process.  
In contrast with this, we directly manipulate shallow features of the pre-trained CLIP visual encoder.} 

\smallskip\noindent\textbf{Prompt Learning.~} Inspired by prompt embedding optimization and automatic prompt generation in natural language processing~\citep{shin2020autoprompt,jiang2020can,zhong2021factual}, ~\citet{zhou2022learning} apply prompt learning for context optimization in VL models, and show that it outperforms prompt engineering in the few-shot setting for different image recognition downstream tasks. CoOp (for Context Optimization)~\citep{zhou2022learning} consists of replacing the prompt context by learnable vectors, which are optimized by minimizing a classification loss using few labeled support images. The weights of both the image and text encoders are frozen during this process. Subsequently, CoCoOp~\citep{zhou2022conditional} adds a lightweight neural network that predicts input-conditional tokens, and improves the performance on unseen classes compared to CoOp. Instead of learning a single prompt, ProDA~\citep{lu2022prompt} estimates the distribution of prompts in the output embedding space, \ie, modeling the classifier weights distribution by a multivariate Gaussian. Such a strategy handles the variability of visual representations.

In this work, the embeddings of the source training images are used to optimize their common concept in the language space. This differs from strategies like~\citep{zhou2022learning} which fix the class name and learn the context in a supervised way using a cross-entropy loss. Our goal is to ``search'' in the text space for a representation that expresses the meta-domain common to the integrity of our source images, \ie, supposing that, globally, all of them share the same semantic class (\ie, driving scenes).

\smallskip\noindent\textbf{Normalization in Deep Learning of CNNs.} To accelerate and stabilize the training of CNNs, Batch Normalization (BN) was proposed in~\citep{ioffe2015batch}. During mini-batch training, BN operation consists of normalizing the batch of features channel-wise and applying a learnable affine transformation. Specifically, BN reads as:
\begin{equation}
	\small
	\texttt{BN}({\mb{f}}) = \bs{\gamma} \left( \frac{\mb{f} - \mu_c(\mb{f})}{\sigma_c(\mb{f})} \right) + \bs{\beta}~,
	\label{eqn:BN}
\end{equation}
where $\mb{f} \in \mathbb{R}^{B \times C \times H \times W}$ and $B,C,H,W$ are respectively the batch size, the number of channels, the height and the width of the feature activation; $\bs{\gamma},\bs{\beta} \in \mathbb{R}^{C}$ are learnable parameters; $\mu_c(.)$ and $\sigma_c(.)$ are two functions that return the mean and standard deviation for each channel, computed across the remaining dimensions of $\mb{f}$. Specifically, for a channel $c$: 
\begin{equation}
	\small
    \mu_c(\mb{f}) = \frac{1}{BHW} \sum_{b=1}^{B} \sum_{h=1}^{H} \sum_{w=1}^{W} \mb{f}_{bchw}~,
	\label{eqn:mu_BN}
\end{equation}

\begin{equation}
	\small
    \sigma_c(\mb{f}) = \sqrt{\frac{1}{BHW} \sum_{b=1}^{B} \sum_{h=1}^{H} \sum_{w=1}^{W} (\mb{f}_{bchw} - \mu_c(\mb{f}))^2 + \epsilon}~.
	\label{eqn:sigma_BN}
\end{equation}
$\epsilon$ is a small positive constant used for numerical stability (\eg, $\epsilon=10^{-5}$). The effectiveness of BN was initially associated to its effect on reducing internal covariate shift, while following research rather argued for its role in smoothing the optimization landscape~\citep{santurkar2018does}. ~\citet{ulyanov2016instance,ulyanov2017improved} show that replacing BN by Instance Normalization (IN) improves image stylization. IN reads as:
\begin{equation}
	\small
	\texttt{IN}({\mb{f}}) = \bs{\gamma} \left( \frac{\mb{f} - \mu(\mb{f})}{\sigma(\mb{f})} \right) + \bs{\beta}~,
	\label{eqn:IN}
\end{equation}
with $\mu(.)$ and $\sigma(.)$ computed across the spatial dimension for each channel $c$ of an instance $b$ in a mini-batch:
\begin{equation}
	\small
    \mu_{bc}(\mb{f}) = \frac{1}{HW} \sum_{h=1}^{H} \sum_{w=1}^{W} \mb{f}_{bchw}~,
	\label{eqn:mu_IN}
\end{equation}
\begin{equation}
	\small
    \sigma_{bc}(\mb{f}) = \sqrt{\frac{1}{HW} \sum_{h=1}^{H} \sum_{w=1}^{W} (\mb{f}_{bchw} - \mu_c(\mb{f}))^2 + \epsilon}~.
	\label{eqn:sigma_IN}
\end{equation}

Leveraging the relation between low-level feature statistics and image style, Adaptive Instance Normalization (AdaIN)~\citep{huang2017arbitrary} transfers style-specific components across features.
In AdaIN, the styles are represented by the channel-wise mean \mbox{$\mu(\mb{f}) \in \mathbb{R}^{C}$} and standard deviation $\sigma(\mb{f}) \in \mathbb{R}^{C}$ of features. Stylizing a source feature $\src{\mb{f}}$ with an arbitrary target style $(\mu(\trg{\mb{f}}), \sigma(\trg{\mb{f}}))$ reads:
\begin{equation}
	\small
	\hspace{-0.2cm}\texttt{AdaIN}(\src{\mb{f}}, \mu(\trg{\mb{f}}), \sigma(\trg{\mb{f}})) = \sigma(\trg{\mb{f}}) \left( \frac{\src{\mb{f}} - \mu(\src{\mb{f}})}{\sigma(\src{\mb{f}})} \right) + \mu(\trg{\mb{f}})~,
	\label{eqn:adain}
\end{equation}
where $\mu(.)$ and $\sigma(.)$ are defined as in~\cref{eqn:mu_IN} and~\cref{eqn:sigma_IN}. Note that there are no learnable parameters in~\cref{eqn:adain}.

These normalization methods, sharing similar mathematical formulations, have all different goals. In this paper, we propose prompt/photo-driven instance normalization (PIN), which is inspired by AdaIN, yet supposes having no access to $\mu(\trg{\mb{f}})$ and $\sigma(\trg{\mb{f}})$.

\section{Domain Adaptation with a single VL Embedding}
\label{sec:DA_VL}
{Our framework, illustrated in \cref{fig:overview}, builds upon CLIP~\citep{radford2021learning}, a Vision-Language model pre-trained on 400M image-text pairs crawled from the Internet.
CLIP trains jointly an image encoder~$\EncI$ and a text encoder~$\EncT$ to learn an expressive representation space that effectively bridges the two modalities. We use such bimodal space in this work to bring image embeddings from a source domain closer to a target domain defined by a single VL embedding, coming either from a textual description (\ie,~zero-shot) or an unlabeled target image~(\ie,~one-shot). This is done by optimizing the style-specific components of the source low-level features, with the final values of these components corresponding to augmented data.

\textbf{For zero-shot}, we explore two types of guidance without access to target data: using either a full textual prompt or a textual prompt combined with an optimized concept.
For the former, we explore the use of simple general prompts describing the target domain, \eg, ``driving at night'' for the night domain. 
Because simple prompts may not be sufficient to represent complex semantics of scenes, we also suggest optimizing a concept $\concept{}$ from the source images, inspired by context optimization~\citep{zhou2022learning}. To form the target prompt, this concept is complemented with our target description, \eg, ``$\concept{}$ at night''.
Our zero-shot domain adaptation framework can interchangeably leverage any of these guidance signals.

\textbf{For one-shot,} we simply replace the single language embedding by the embedding of a single unlabeled target image. The optimization thus operates in the image embedding space, \ie, the space defined by the image encoder. We later show that this setting might be useful when the target domain conditions are nuanced and not trivial to verbalize.

Overall, our goal is to optimize affine transformations of low-level source features so that the embeddings of these features become closer to their imaginary counterparts in the targeted domain (\cref{fig:overview}, left),
while crucially preserving their semantic content. 
The learned augmentations can then be applied to synthesize, in a zero-shot/one-shot fashion, features that correspond to the target domain and can subsequently be used to fine-tune the model (\cref{fig:overview}, middle). This ultimately allows inference on domains only described by either a single prompt or an image at training time (\cref{fig:overview}, right).

{Our approach faces several challenges:
\textit{(i)}~How to ``mine'' style information relative to a target domain using only one VL embedding? 
\textit{(ii)}~How to preserve pixel-wise semantics (\ie, content) during style mining?
\textit{(iii)}~Based on such mined styles, how to adapt the source model to the target domain?
We address these questions in the following.}

\smallskip\noindent\textbf{Problem formulation.~}
Our main task is semantic segmentation, that is, pixel-wise classification of input images into semantic segments.
We start from a $K$-class segmentation model $M$, trained on a labeled source domain dataset \makebox{$\src{\mc{D}}=\{(\src{\mb{I}}, \src{\mb{y}}) \mid \src{\mb{I}} \in\mathbb{R}^{H\times W \times 3}, \src{\mb{y}} \in \{0,1\}^{H\times W \times K}\}$}, where $\src{\mb{I}}$ represents an image and $\src{\mb{y}}$ its ground-truth annotation.
The segmenter $M$ is composed of a CLIP image encoder~$\EncI$ (\eg, ResNet-50) as the frozen feature extractor backbone $M_\text{feat}$ and a randomly initialized pixel classification head $M_\text{cls}$:  $M=(M_\text{feat}, M_\text{cls})$.
Our goal is to adapt the model $M$ such that its performance on a test target dataset \makebox{$\trg{\mc{D}}=\{\trg{\mb{I}} \mid \trg{\mb{I}} \in\mathbb{R}^{H\times W \times 3}\}$} is improved, provided solely a single VL latent embedding as information about the target domain. 
This is enabled by access to a target embedding, $\targetFeat{}$, which is derived either from a prompt using the frozen text encoder, or from a single unlabeled target image using the frozen image encoder.

We train $M$  in a supervised manner for the semantic segmentation task on the source domain.
Interestingly, we empirically show in \cref{tab:baselines} that keeping the feature extractor $M_\text{feat}$ frozen helps alleviate the overfitting risk to the source domain in favor of generalization; not a universal finding, this observation is limited to our experimental setting specified in~\cref{sec:exp_details}. To minimize the interference of generalization effects caused by different training strategies in the adaptation results, we systematically freeze $M_\text{feat}$ when training on source and later fine-tuning for adaptation.

From the extractor, we remove the attention pooling head of $\EncI$ to keep the spatial information for the pixel classifier.
We denote $\mb{f}$ the intermediate feature activations extracted by $M_\text{feat}$ and $\emb{\mb{f}}$ their corresponding CLIP embeddings.
In \cref{fig:method} we illustrate the difference between $\mb{f}$ and $\emb{\mb{f}}$.}

\begin{table}[t]
	\setlength{\tabcolsep}{0.03\linewidth}
	\centering
	\begin{tabular}{cccccc}
		\toprule
		$M_\text{feat}$\,\freeze & CS &  Night &  Snow &  Rain & GTA5 \\
		\midrule
		Yes & 66.82 & \textbf{18.31} & \textbf{39.28} & \textbf{38.20} & \textbf{39.59} \\
		No & \textbf{69.17} & 14.40 & 22.27 & 26.33 & 32.91\\
		\bottomrule
	\end{tabular}
    \smallskip\caption{
    \small\textbf{Segmentation performance of source-only models.} Mean Intersection-over-Union (mIoU\,\%) is reported on the ``night'', ``snow'', and ``rain'' subsets of the ACDC~\citep{sakaridis2021acdc} \textit{validation set}, as well as on a subset of $1\,000$ images from GTA5. Models are trained on Cityscapes (`CS'). `$M_\text{feat}$\texttwemoji{2744}' indicates a frozen backbone.}
\label{tab:baselines}
\end{table}

\smallskip\noindent\textbf{Overview of the proposed method.~} 
{Our solution is to 
mine styles (\ie channel-wise statistics of the features) using source-domain low-level features set $\src{\mc{F}}{=}\{\src{\mb{f}} {\mid} \src{\mb{f}} {=} \texttt{feat-ext}(M_\text{feat}, \src{\mb{I}})\}$ and $\targetFeat$.
For generality, $\texttt{feat-ext}(\cdot)$ can pull 
features from any desired layer. However, we later show that using the features from the earliest layers yields the best results.}

{The $\texttt{augment}(\cdot)$ operation, depicted in \cref{fig:method}, augments the style-specific components of $\src{\mb{f}}$ with guidance from $\targetFeat$, synthesizing $\stot{\mb{f}}$ with 
style information from the target domain.
We emphasize that 
the features $\src{\mb{f}}$ and $\stot{\mb{f}}$ have the same size 
and identical semantic content, though they encapsulate different visual styles.
For adaptation, the source features $\src{\mb{f}}$ are augmented with the mined styles then used to
fine-tune the classifier $M_\text{cls}$, leading to the final adapted model.} 

\subsection{Instance Normalization guided by single VL embedding}
\label{sec:IN_VL}

{We draw inspiration from Adaptive Instance Normalization (AdaIN)~\citep{huang2017arbitrary}, an elegant formulation for transferring style-specific components across features. AdaIN is defined in~\cref{eqn:adain}.}

{We design our augmentation strategy 
around AdaIN as it can effectively manipulate the style information with a small set of parameters. In the following, we present our 
augmentation strategy which 
mines target styles.}

\subsubsection{Prompt-driven Instance Normalization}
\label{sec:PIN_prompt}
In the zero-shot setting, we suppose having no access to any image from the target domain. 
Consequently, $\mu(\trg{\mb{f}})$ and $\sigma(\trg{\mb{f}})$ in~\cref{eqn:adain} are unknown. Yet, we suppose having a single text condition in natural language (\ie, a prompt~$\prompt{}$) describing the target domain. Thus, we propose \makebox{Prompt-driven} Instance Normalization (PIN) as
\begin{equation}
	\small
	\texttt{PIN}_{\targetFeat{}}(\src{\mb{f}}, \bs{\mu}, \bs{\sigma}) := \bs{\sigma} \left( \frac{\src{\mb{f}} - \mu(\src{\mb{f}})}{\sigma(\src{\mb{f}})} \right) + \bs{\mu}~,
	\label{eqn:PIN}
\end{equation}
where ${\bs{\mu}} \in \mathbb{R}^C$ and ${\bs{\sigma} \in \mathbb{R}^C}$ are optimizable variables driven by $\targetFeat= \EncT(\prompt{})$.

{We aim to augment source image features $\src{\mc{F}}$ such that they capture the style of the unseen target domain.
Here, the prompt describing a target domain could be fairly generic.
For instance, one can use prompts like
``driving at night'' or ``driving under rain'' 
to bring source features closer to the nighttime or rainy domains.

\subsubsection{\textbf{Concept optimization}}
\label{sec:concept_learning}
In domain adaptation, despite the discrepancies between source and target data, they still share common characteristics (\eg, ``driving'' data). 
Interestingly, our experiments show that prompt-based adaptation performs better when the prompt includes a word describing the common characteristics (\eg, the word ``\textit{driving}'' in ``\textit{driving} at night'', ``\textit{driving} under rain'', \etc), though this improvement comes at the cost of prompt engineering.
To prevent this, we draw inspiration from CoOp~\citep{zhou2022learning} and seek to optimize a word embedding describing the common characteristics. 
While similar in spirit, our approach differs from CoOp in methodology and implementation as the latter requires access to target data and their labels.
Instead, we optimize a concept using only \textit{source} images and a textual style description as guidance (\eg, ``in clear weather'').

In practice, we apply concept optimization in two steps. In the first step, 
we construct a prompt $\prompt{}$ in the form of \texttt{<concept>} + ``in clear weather'', where \texttt{<concept>} represents optimizable parameters in the word embedding space. 
The optimization is done using SGD with the following objective: 
\begin{equation}
	\small
    \mc{L}_{<\texttt{concept}>} = 1 - \frac{\EncT(\prompt{}).\EncI(\src{\mb{I}})}{\|\EncT(\prompt{})\| \|\EncI(\src{\mb{I}})\|}~.
    \label{eqn:concept_learning}
\end{equation}
The word embedding is optimized by minimizing the cosine similarity between the representations of the source images and the prompt representation. The final value of \texttt{<concept>} is denoted $\concept{}$.
In the second step, the target prompt $\promptConcept$ is constructed by concatenating $\concept{}$ with target-specific texts, such as ``$\concept{}$ at night'' or ``$\concept{}$ under rain''.
Our proposed \textit{concept optimization} is illustrated in~\cref{fig:concept_optim}.
We then apply PIN (\cref{eqn:PIN}), though using as $\targetFeat$ the VL embedding of $\promptConcept$, \ie, $\targetFeat=\EncT(\promptConcept)$.

\begin{figure}
	\centering
	\includegraphics[trim=2 0 0 0, clip,width=1\linewidth]{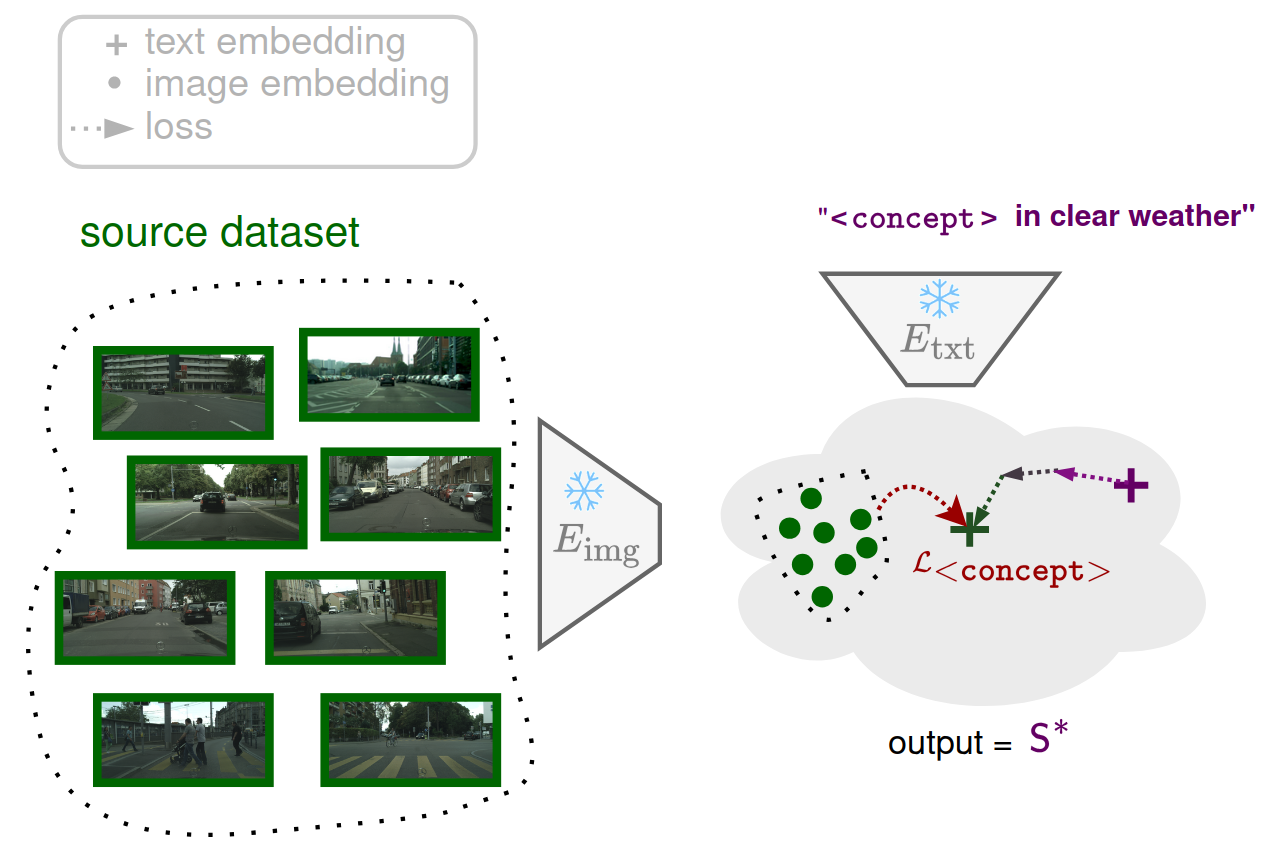}
	\caption{
    \small\textbf{Concept optimization.} A learnable \texttt{<concept>} token is optimized in the word embedding space using a cosine distance objective $\color{darkred}\mc{L}_{\texttt{<concept>}}$, encouraging the text embedding to align closely with embeddings of source images. The final optimized word embedding is denoted by $\mathsf{S}^*$.
 }
\label{fig:concept_optim}
\end{figure}

\subsubsection{Photo-driven Instance Normalization}
\label{sec:photodrivennorm}

Within our framework, $\targetFeat$ is simply an embedding guiding the optimization of $\bs{\mu}$ and $\bs{\sigma}$. 
While we have explored the use of text to obtain $\targetFeat$, certain peculiar conditions may be challenging to describe this way.
Alternatively, our framework also allows getting the VL embedding from a single unlabeled target image $\trg{\mb{I}}$.

In such case, PIN (\cref{eqn:PIN}) is applied using the embedding of the target image (\ie,~$\targetFeat=\EncI(\trg{\mb{I}})$) as guidance. In this setting, PIN refers to \textit{Photo-driven Instance Normalization}.

\subsection{Style mining with PIN}
\label{sec:stylemining}

In the sections above, we described how PIN can leverage either a target prompt (\cref{sec:PIN_prompt}), the combination of a source concept and a~target prompt (\cref{sec:concept_learning}), or a target image~(\cref{sec:photodrivennorm}). We now detail how to mine a collection of styles.

\begin{figure}
	\centering
	\includegraphics[trim=0 0 35cm 0, clip,width=0.95\linewidth]{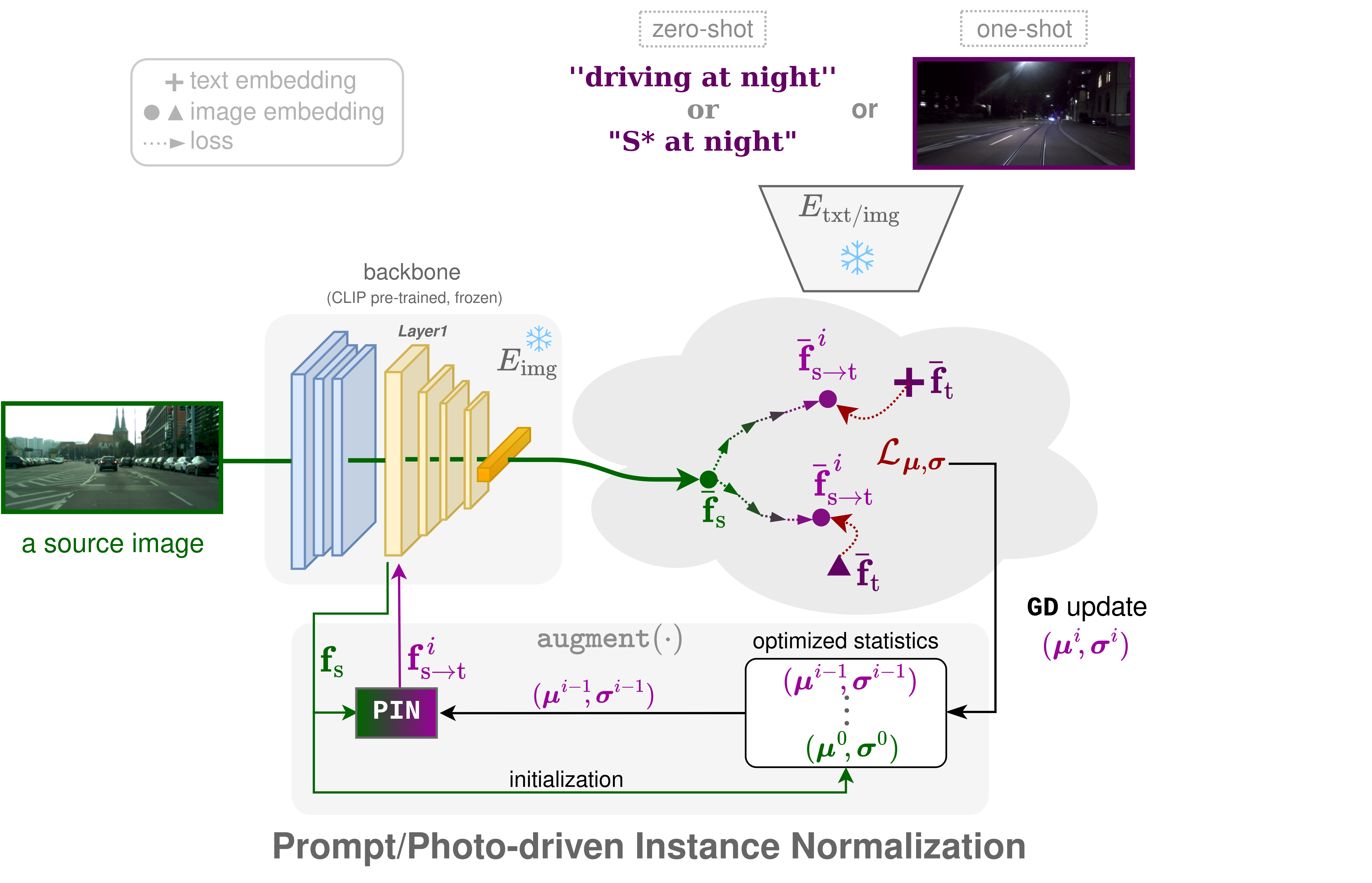}
	\caption{
 \small
    \textbf{Target style mining from a source image.} This figure illustrates the optimization loop described in \cref{algo:style_mining}. A source image is passed through the CLIP image encoder $\EncI$ to extract low-level features $\color{srccolor} \src{\mb{f}}$ and the corresponding CLIP embedding $\color{srccolor} \src{\emb{\mb{f}}}$. At each optimization step $i$, $\texttt{augment}(\cdot)$ injects the previous style parameters $\color{intermcolor}(\bs{\mu}^{i-1}, \bs{\sigma}^{i-1})$ into the features $\color{srccolor} \src{\mb{f}}$ using the \texttt{PIN} layer, producing the stylized features $\color{intermcolor} \textbf{f}^i_{\text{s} \shortrightarrow \text{t}}$ and their embedding $\color{intermcolor} \bar{\textbf{f}}^i_{\text{s} \shortrightarrow \text{t}}$. The loss $\color{darkred}\mc{L}_{\bs{\mu}, \bs{\sigma}}$ is computed as the cosine distance between~$\color{intermcolor}\bar{\textbf{f}}^i_{\text{s}\shortrightarrow\text{t}}$ and the target embedding~\textcolor{trgcolor}{$\protect\targetFeat$}, which can come from a text prompt (\eg, ``driving at night''), a partially optimized prompt (\eg, ``$\mathsf{S}^*$ at night''), or an unlabeled target image. Gradients from this loss update the style parameters to $\textcolor{intermcolor}{(\bs{\mu}^{i}, \bs{\sigma}^{i})}$ via gradient descent.
    }
\label{fig:method}
\end{figure}

\begin{algorithm}[t]
	\scriptsize
    \SetAlgoLined	
	\SetKwFunction{Mean}{mean}
	\SetKwFunction{Std}{std}
    \SetKwFunction{PIN}{PIN}
	\SetKwFunction{Ge}{get-embedding}
	\SetKwInOut{Input}{Input} 
	\SetKwInOut{Output}{Output} 
	\SetKwInOut{Parameter}{Param}
	
	\Input{Set $\src{\mc{F}}$ of source image features.\\
		VL embedding $\targetFeat$.\\
	}
	\Parameter{Number $N$ of optimization steps.\\
		Learning rate $lr$ and momentum $m$\\
            ~of gradient descend (\texttt{GD}).}
	\Output{Set $\stot{\mc{S}}$ of target styles.} 
	
	$\stot{\mc{S}} \gets \emptyset$ \\
	\ForEach{$\src{\mb{f}} \in \src{\mc{F}}$}{%
		$\bs{\mu}^{0} \gets \Mean(\src{\mb{f}})$
  
		$\bs{\sigma}^{0} \gets \Std(\src{\mb{f}})$
  
		\tcp{Optimization}
		\For{$\text{i} = 1,2, \cdots, N$} {%
			$\mb{f}_{\text{s}\shortrightarrow\text{t}}^{i} \gets \PIN_{\targetFeat}(\src{\mb{f}}, \bs{\mu}^{i-1}, \bs{\sigma}^{i-1})$
			
            $\emb{\mb{f}}_{\text{s}\shortrightarrow\text{t}}^{i} \gets \Ge(\mb{f}_{\text{s}\shortrightarrow\text{t}}^{i})$
            
			\mbox{$\bs{\mu}^{i} \gets \texttt{GD}^{lr}_{m}(\bs{\mu}^{i-1}, \nabla_{\bs{\mu}}\mc{L}_{\bs{\mu}, \bs{\sigma}}(\emb{\mb{f}}_{\text{s}\shortrightarrow\text{t}}^{i}, \targetFeat))$}
   
			\mbox{$\bs{\sigma}^{i} \gets \texttt{GD}^{lr}_{m}(\bs{\sigma}^{i-1}, \nabla_{\bs{\sigma}}\mc{L}_{\bs{\mu}, \bs{\sigma}}(\emb{\mb{f}}_{\text{s}\shortrightarrow\text{t}}^{i}, \targetFeat))$}
		}
    	   $(\bs{\mu}_{t}, \bs{\sigma}_{t}) \gets (\bs{\mu}^{N}, \bs{\sigma}^{N})$ \\
		$\stot{\mc{S}} \gets \stot{\mc{S}} \cup \{(\bs{\mu}_\text{t}, \bs{\sigma}_\text{t})\}$
	}
	\caption{
		Style Mining (see Fig.~\protect{\ref{fig:method}})}
	\label{algo:style_mining}
\end{algorithm}

{We describe in~\cref{algo:style_mining} the first step of our feature augmentation procedure: mining the set $\stot{\mc{S}}$ of 
styles in targeted domain.
For each source feature activation $\src{\mb{f}} \in \src{\mc{F}}$, we want to mine style statistics corresponding to an imaginary target feature activation $\trg{\mb{f}}$. 
To this end, we formulate style mining as an optimization problem over the original source feature $\src{\mb{f}}$, \ie, 
optimizing $(\bs{\mu}, \bs{\sigma})$ in~\cref{eqn:PIN}. 
The optimization objective is defined as the cosine distance 
in the CLIP latent space between the CLIP embedding $\emb{\mb{f}}_{\text{s}\shortrightarrow\text{t}}$ of the stylized feature 
\mbox{$\stot{\mb{f}} = \texttt{PIN}(\src{\mb{f}}, \bs{\mu}, \bs{\sigma})$} and the VL embedding $\targetFeat$ of target domain:
\begin{equation}
	\small
	\mc{L}_{\bs{\mu}, \bs{\sigma}}(\emb{\mb{f}}_{\text{s}\shortrightarrow\text{t}}, \targetFeat) = 1-\frac{\emb{\mb{f}}_{\text{s}\shortrightarrow\text{t}}\cdot\targetFeat}{\|\emb{\mb{f}}_{\text{s}\shortrightarrow\text{t}}\|\,\|\targetFeat\|}\,.
	\label{eqn:loss_dir}
\end{equation}
This CLIP-space cosine distance, already used in prior text-driven image editing works~\citep{patashnik2021styleclip}, 
aims to steer the stylized features in the direction of the target VL embedding. 
One step of the optimization is illustrated in \cref{fig:method}. 
In practice, we run several such steps per image~(\ie,~instance) leading to the mined target style denoted $(\bs{\mu}_\text{t}, \bs{\sigma}_\text{t})$.}

{As there might be a variety of styles in a target domain, our mining populates the $\stot{\mc{S}}$ set with as many variations of target style as there are source images, hence
$|\stot{\mc{S}}| = |\src{\mc{D}}|$.}

{Intuitively, our simple augmentation strategy can be seen as a cost-efficient way to cover the distribution of the target domain by starting from different anchor points in CLIP latent space coming from the source images and steering them in the direction of the target VL embedding.
This mitigates the diversity problem discussed in one-shot feature augmentation~\citep{luo2020adversarial,wu2022style}.}

\begin{algorithm}[t!]
	\small
	\SetAlgoLined
	\SetKwFunction{Train}{train}
	\SetKwFunction{Fn}{fine-tune}
	\SetKwFunction{Feat}{feat-ext}
	\SetKwFunction{Init}{init}
	\SetKwFunction{Mine}{style-mining}
	\SetKwInOut{Input}{Input}  
	\Input{Source dataset $\src{\mc{D}}=\{(\src{\mb{I}}, \src{\mb{y}})\}$.\\
		CLIP encoders $\EncI$ and $\EncT$.\\
		  Target VL embedding $\targetFeat$.\\
		Feature backbone $M_\text{feat} \xleftarrow{} \EncI$.\\
		Source model: $M = (M_\text{feat}, M_\text{cls})$.\\
	}
	\KwResult{\mbox{Target-adapted model $M' = (M_\text{feat}, M'_\text{cls})$}.}
	\tcp{Initialization}
	$M \gets \Train(M, \src{\mc{D}})$ 
 
	\tcp{Feature Augmentation}
	$\src{\mc{F}} \gets \Feat(M_\text{feat}, \{\src{\mb{I}}\})$
	
    $\stot{\mc{S}} \gets \texttt{augment}(\src{\mc{F}}, \targetFeat)$ 
    
    \tcp{Adaptation}
	$M' \gets \Fn(M, \src{\mc{F}}, \stot{\mc{S}}, \{\src{\mb{y}}\})$ 
    \caption{VL-guided UDA}
	\label{algo:PODA_overall}
\end{algorithm}

\subsection{Fine-tuning for Adaptation}
\label{sec:fine_tune}
{For adaptation, at each training iteration, we stylize the source features using a mined target style $({\bs{\mu}}_\text{t}, \bs{\sigma}_\text{t})$ randomly selected from $\stot{\mc{S}}$. The augmented features are computed as 
\makebox{$\mb{f}_{\text{s}\shortrightarrow\text{t}} = 
 \texttt{AdaIN}(\src{\mb{f}}, \bs{\mu}_\text{t}, \bs{\sigma}_\text{t})$ }
and are used for fine-tuning the classifier $M_\text{cls}$ of the segmenter $M$ (\cref{fig:overview}, middle). 
As we only adjust the feature style, which keeps the semantic-content unchanged~\citep{huang2017arbitrary}, we can still use the labels ${\src{\mb{y}}}$ to train the classifier in a supervised manner with a standard segmentation loss (\ie, cross-entropy).
To this end, we simply forward augmented features through remaining layers in $M_\text{feat}$ followed by $M_\text{cls}$.
In the backward pass, only weights of $M_\text{cls}$ are updated by the loss gradients.
We denote the fine-tuned model as $M' = (M_\text{feat}, M'_\text{cls})$. 

\cref{algo:PODA_overall} presents the high-level pseudo-code for VL-guided adaptation: from source-only training as model initialization, to VL embedding-driven feature augmentation, to zero-shot/one-shot model adaptation.

In \method{} and \methodconcept{}, $M'$ is evaluated on images with conditions and styles that were never seen during any of the training stages yet described by a single prompt, while in \methodone{} it is evaluated on the validation set of the same dataset from which the target image was sampled.

}

\section{Experiments}

\subsection{Implementation details}
\label{sec:exp_details}
{We use the DeepLabv3+ architecture~\citep{chen2018encoder} with the backbone $M_\text{feat}$ initialized from the image encoder $\EncI$ of the pre-trained CLIP-ResNet-50 model.\footnote{\url{https://github.com/openai/CLIP}}}

\smallskip\noindent\textbf{Source-only training.~}
{The network is trained for  
$200$k iterations on random $768 \times 768$ crops with a batch size of~$2$.
We use a polynomial learning rate schedule with initial $lr{=}10^{-1}$ for the classifier {and $lr{=}10^{-4}$ for backbone when it is not frozen} (see~\cref{tab:baselines}). 
Optimization is done with Stochastic Gradient Descent (SGD)~\citep{bottou2010large}, momentum $0.9$ and weight decay $10^{-4}$.
{We apply standard color jittering and horizontal flip to crops.}}

\smallskip\noindent\textbf{Concept optimization.~} $\concept{}$ is optimized from the source images using SGD for $10$ epochs, with a batch size of~16. The learning rate is set to $10^{-4}$.

\smallskip\noindent\textbf{PIN optimization.~}
For the style mining step, we use the source feature activations after the first layer (\textit{Layer1}):  $\src{\mb{f}}\in \mathbb{R}^{192 \times 192 \times 256}$.
The style parameters $\bs{\mu}$ and $\bs{\sigma}$ are $256$-dim real vectors.
The CLIP embeddings are $1024$-dim vectors for \mbox{CLIP-ResNet-50} backbone, and $512$-dim vectors for \mbox{CLIP-ResNet-101}.
We adopt the Imagenet templates from~\citep{radford2021learning} to encode the target descriptions $\prompt$. For \method{}, \methodconcept{} and \methodone, PIN is optimized for $100$ iterations using Gradient Descent (GD) on batches of 16 feature instances with a learning rate $lr$=1.0.

\smallskip\noindent\textbf{Classifier fine-tuning.~} 
{Starting from the source-only trained model, we fine-tune the classifier $M_\text{cls}$ on batches of $8$~augmented features $\stot{\mb{f}}$ 
for $2,000$ iterations.
Polynomial schedule is used with the initial $lr=10^{-2}$. We always use the last checkpoint for evaluation.}

\smallskip\noindent\textbf{Datasets.~} 
{As source, we use Cityscapes~\citep{cordts2016cityscapes}, composed of $2,975$ training and $500$ validation images featuring $19$ semantic classes. 
Though we adapt towards a prompt or an image \textit{not} a dataset, we need ad-hoc datasets to test on.
We report main results using ACDC~\citep{sakaridis2021acdc} because it has urban images captured in adverse conditions.
We also study the applicability of our method to the two settings of
\DAsetting{real}{synthetic} (Cityscapes as source, and evaluating on GTA5~\citep{richter2016playing}) and \DAsetting{synthetic}{real} (GTA5 as source, and evaluating on Cityscapes).
We evaluate on the validation set when provided, and for GTA5 evaluation we use a random subset of $1,000$ images.}

\smallskip\noindent\textbf{Evaluation protocol.~} 
{Mean Intersection over Union (mIoU\%) is used to measure adaptation performance. We test all models on target images at their original resolutions.
We always report the mean and standard deviation over five models trained with different random seeds.} 

\label{sec:main_res}

\subsection{\method}
{We consider the following adaptation scenarios: \DAsetting{day}{night},~\DAsetting{clear}{snow},~\DAsetting{clear}{rain},
~\DAsetting{real}{synthetic} and~\DAsetting{synthetic}{real}.
We report zero-shot adaptation results of~\method~in the addressed set-ups, comparing against two main baselines: CLIPstyler~\citep{kwon2022clipstyler} for zero-shot style transfer and SM-PPM~\citep{wu2022style} for one-shot UDA.
Both~\method~and CLIPstyler models do not see any target images during training.
{In this study, we arbitrarily choose a simple prompt to describe each domain.
We show later in~\cref{sec:abla} more results using other relevant prompts with similar meanings -- showing
that our adaptation gain exhibits only a slight sensitivity to prompt selection.}
For SM-PPM, one random target image from the training set is used.}

\smallskip\noindent\textbf{Comparison to CLIPstyler.~}
{CLIPstyler~\citep{kwon2022clipstyler} is a style transfer method that {also} makes use of the pre-trained CLIP model {but} for zero-shot stylizing of source images.
We consider CLIPstyler\footnote{We use official code \url{https://github.com/cyclomon/CLIPstyler} and follow the recommended configs.} as the most comparable zero-shot baseline for \method~as both are built upon CLIP, though with different mechanisms and different objectives.
Designed for style transfer, CLIPstyler produces images that exhibit characteristic styles of the input text prompt.
However the stylized images 
{can have multiple} artifacts which hinder their usability in the downstream segmentation task.
This is visible in~\cref{fig:clipstyler} which shows stylized examples from CLIPstyler with \method target prompts. 
{Zooming in, we note that stylization of snow or game added snowy roads or Atari game \textit{on the buildings}, respectively.}}

\begin{figure}[h]
	\newcommand{\rowqual}[1]{ %
	    \includegraphics[width=0.19\linewidth]{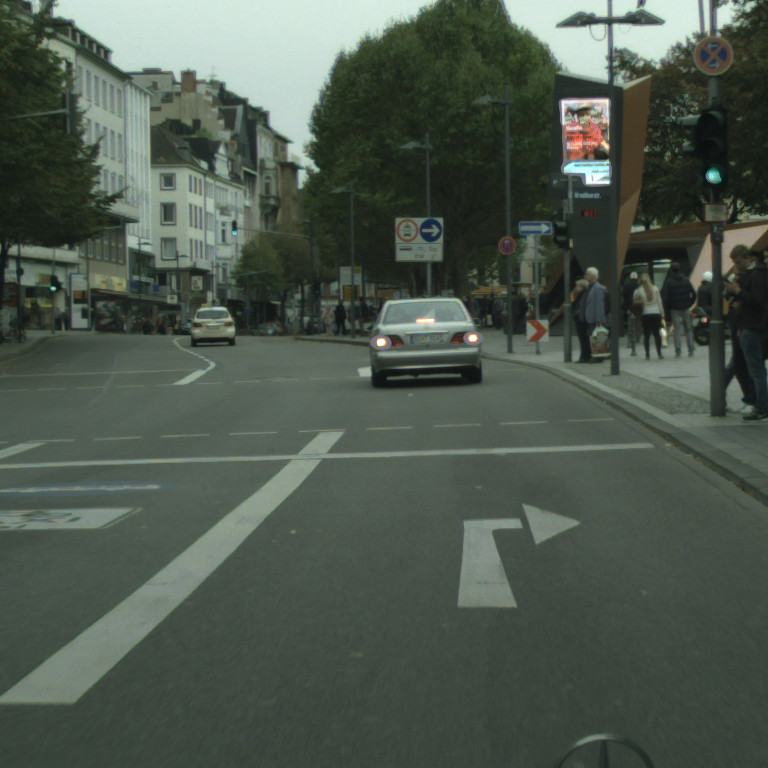}&
		\includegraphics[width=0.19\linewidth]{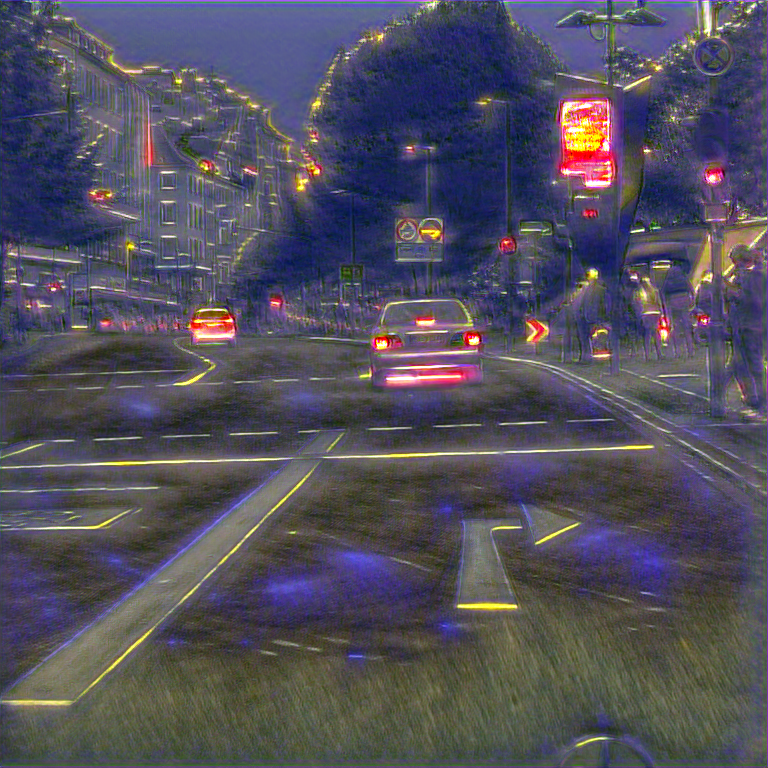}&
		\includegraphics[width=0.19\linewidth]{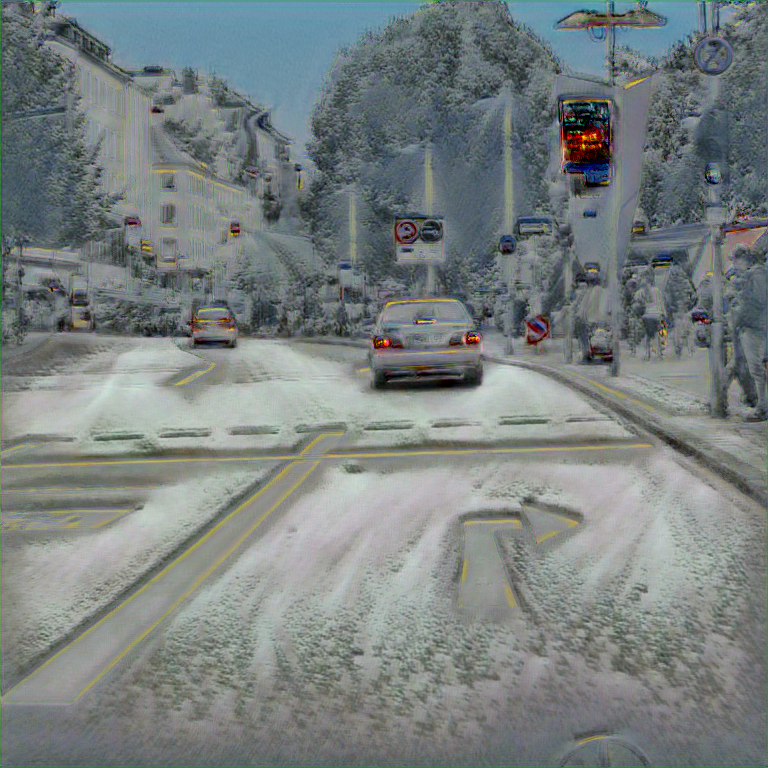}&
		\includegraphics[width=0.19\linewidth]{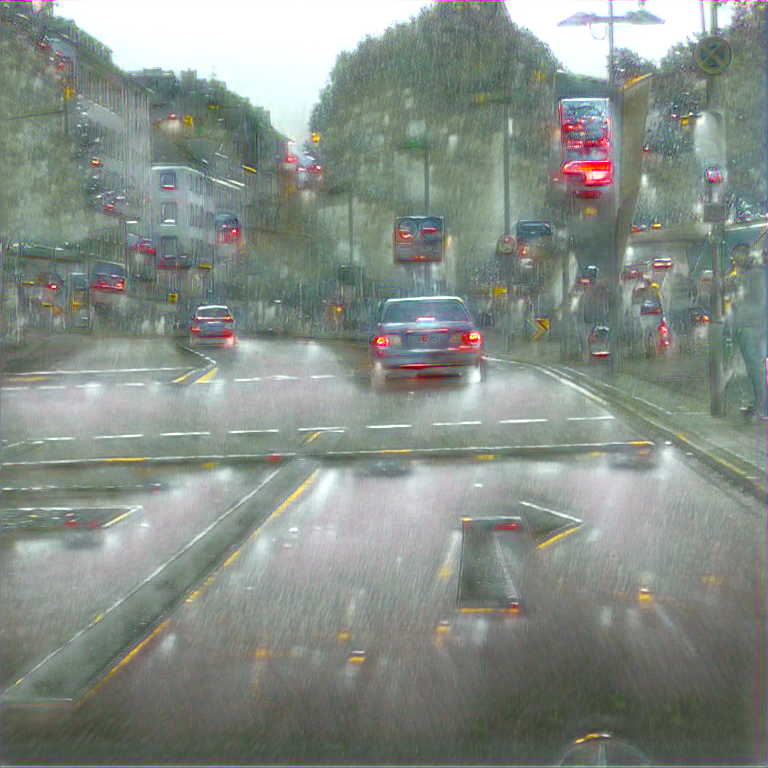}&
		\includegraphics[width=0.19\linewidth]{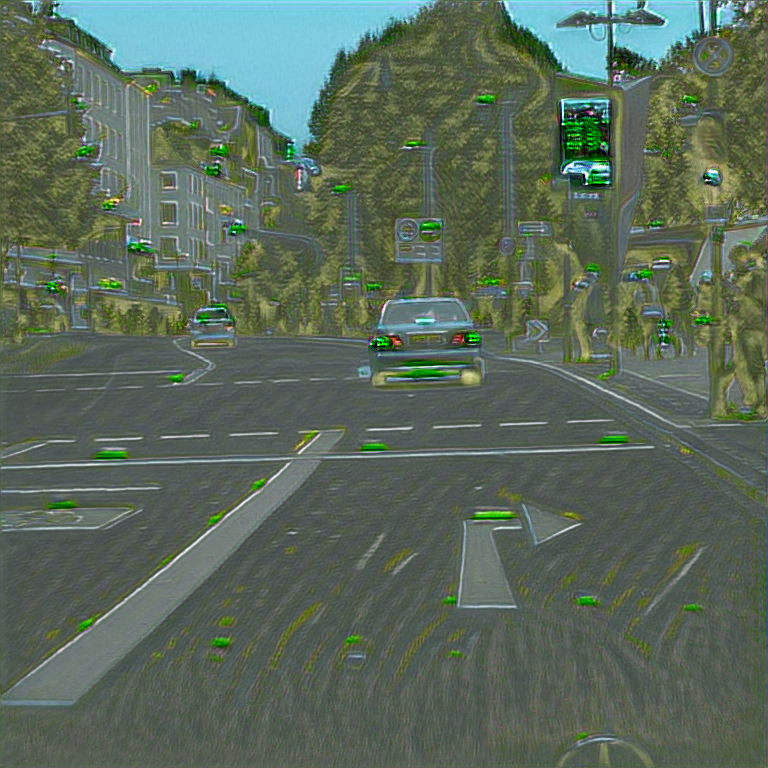}
	}
	\small	
	\setlength{\tabcolsep}{0.002\linewidth}
	\centering
	\begin{tabular}{cccccc}
				& Cityscapes & night & snow & rain & game\\
		\rotatebox{90}{} & \rowqual{CLIPstyler_visuals/33}\\ 
	\end{tabular}
    \smallskip\caption{
    \small\textbf{CLIPstyler~\citep{kwon2022clipstyler} stylization.} An example Cityscapes image stylized using ad-hoc target prompts. The resulting translated images exhibit visible artifacts, which may negatively affect domain adaptation performance, \eg, in the rain scenario shown in~\cref{tab:main_results}.
    }
    \label{fig:clipstyler}
\end{figure}

{Starting from source-only model, we fine-tune the classifier on stylized images, as similarly done in \method{} {with the augmented features}. Table \ref{tab:main_results} compares \method~against the source-only model and CLIPstyler.~\method~consistently {outperforms the two baselines.}
CLIPstyler brings
some improvements over source-only in~\DAsetting{Cityscapes}{Night} and \DAsetting{Cityscapes}{Snow}. 
In other scenarios, \eg, rain, {CLIPstyler even performs worse than source-only.}}

\begin{table}[t!]
		\setlength{\tabcolsep}{0.04\linewidth}
		\centering
		\resizebox{0.98\linewidth}{!}{
                \newcolumntype{H}{>{\setbox0=\hbox\bgroup}c<{\egroup}@{}}
			\begin{tabular}{lllc}
			\toprule
			Source
			& Target eval.
			& Method
			& mIoU[\%]\\
			\midrule
        \multirow{16}{*}{CS}& \multicolumn{3}{c}{\cellcolor{gray!34}$\prompt$ = ``driving at night''}\\
			& \multirow{3}{*}{ACDC Night} & source-only & 18.31\\
    & & CLIPstyler & 21.38\vartn{0.36} \\
    & & \method & \textbf{25.03}\vartn{0.48} \\
        & \multicolumn{3}{c}{\cellcolor{gray!34}$\prompt$ = ``driving in snow''}\\
        & \multirow{3}{*}{ACDC Snow} & source-only &  39.28\\
    & & CLIPstyler & 41.09\vartn{0.17} \\
        & & \method & \textbf{43.90}\vartn{0.53} \\
        &\multicolumn{3}{c}{\cellcolor{gray!34}$\prompt$ = ``driving under rain''}\\
		& \multirow{3}{*}{ACDC Rain} & source-only & 38.20\\
  & & CLIPstyler & 37.17\vartn{0.10} \\
  & & \method &  \textbf{42.31}\vartn{0.55}\\
    &\multicolumn{3}{c}{\cellcolor{gray!34}$\prompt$ = ``driving in a game''}\\
    & \multirow{3}{*}{GTA5} & source-only & 39.59\\
    & & CLIPstyler & 38.73\vartn{0.16} \\
    & & \method & \textbf{41.07}\vartn{0.48}\\
	\arrayrulecolor{black}		
	\midrule
	\multirow{4}{*}{GTA5} &\multicolumn{3}{c}{\cellcolor{gray!34}$\prompt$ = ``driving''}\\
    & \multirow{3}{*}{CS} & source-only &  36.38 \\
    & & CLIPstyler &  31.50\vartn{0.21} \\
    & & \method & \textbf{40.08}\vartn{0.52} \\
    \bottomrule
    \end{tabular}}
    \smallskip\caption{
    \small\textbf{Zero-shot domain adaptation for semantic segmentation.} We report mIoU (\%) performance of~\method~compared to CLIPstyler~\citep{kwon2022clipstyler} and the source-only baseline. Results are grouped by source domain and target prompts~({$\prompt$}). `CS' denotes Cityscapes dataset~\citep{cordts2016cityscapes}.
    }
	\label{tab:main_results}
\end{table}

{\DAsetting{Real}{synthetic} is an interesting though under-explored adaptation scenario. 
One potential application of \DAsetting{real}{synthetic} is for model validation in the industry, where some hazardous validations like driving accidents must be done in the virtual space.
Here we test if our zero-shot mechanism can be also applied to this particular setting.
Similarly, \method~outperforms both baselines.
Also in the reverse~\DAsetting{synthetic}{real} setting, again our method performs the best.
CLIPstyler undergoes almost $5\%$ drops in mIoU compared to source-only.}

We note that \method{} does not yield equal improvements over the source-only model across all domains, nor does CLIPstyler. This phenomenon is common in domain adaptation, where the effectiveness might depend on multiple factors, including the architecture, the pretraining strategy, the source domain, the target domain, and the adaptation method itself. This observation holds in unsupervised domain adaptation~\citep{zou2018unsupervised,hoyer2022daformer,ganin2015unsupervised,kang2019contrastive,xiao2021dynamic} and domain generalization~\citep{zhoudomain,wang2021learning,li2021simple,choi2021robustnet,lee2022wildnet}.

Domains like night exhibit a larger gap from the source domain (\ie, day and clear weather) due to significant illumination variations and increased noise. This gap is reflected in the low performance of
source-only model on night images (\cf \cref{tab:main_results}). Our method is particularly effective at addressing this domain, leading to a performance gain of $\approx+7$ mIoU over the source-only model. This improvement is enabled by the use of a simple prompt such as ``driving at night'', which helps bridge the
gap by aligning low-level feature statistics. In contrast, domains like snow and rain, which share more visual cues with the source domain, exhibit a smaller domain gap, resulting in $\approx+4\%$ mIoU gain.

{We argue on the simplicity of our method that only introduces minimal changes to the feature statistics, yet such changes are crucial for target adaptation.
CLIPstyler, designed for style transfer, involves training an additional StyleNet with $\approx{}615$k parameters for synthesizing the stylized images.
We draw on the simplicity of~\method~to explain why it is more advantageous than CLIPstyler for downstream tasks like semantic segmentation: the minimal statistics changes help avoiding significant drifts on the feature manifold which may otherwise result in unwanted errors.

\smallskip\noindent\textbf{Computation time.~} We measure the average computation time of \method{} using a \mbox{ResNet-50} backbone and a DeepLabV3+ decoder on an RTX 2080TI GPU, showing that our method incurs a minimal overhead. In details, style mining with PIN (\cref{sec:stylemining}) takes $0.3$ second per instance, being $200\times$ faster than CLIPStyler, and resulting in \mbox{$\approx15$} minutes for the whole Cityscapes dataset. Fine-tuning for adaptation (\cref{sec:fine_tune}) with our stylization takes 14 minutes for the 2k iterations -- being 30 seconds longer than Empirical Risk Minimization (ERM). 
Finally, our method adds no computational overhead at inference time, as it does not introduce any additional parameters. Inference on a full-resolution ACDC image takes \mbox{$\approx0.11$} seconds on average.

\smallskip\noindent\textbf{Qualitative results.~}{We show in \cref{fig:qualres} qualitative examples of predictions from source-only and \method~models.}

\begin{figure}[t!]
    \newcommand{\rowqual}[1]{ %
        \includegraphics[width=0.55\linewidth]{#1image.png}&%
        \includegraphics[width=0.55\linewidth]{#1target.png}&%
        \includegraphics[width=0.55\linewidth]{#1pred_src.png}&%
        \includegraphics[width=0.55\linewidth]{#1pred_ours.png}
    }
    \setlength{\tabcolsep}{0.002\linewidth}
    \centering
    \newcolumntype{H}{>{\setbox0=\hbox\bgroup}c<{\egroup}@{}}
    \resizebox{1.0\linewidth}{!}{%
    \begin{tabular}{ccccc}
                & {Input} & {Ground Truth} & {Source-only} & {\method}\\
        \multirow{1}{*}[4.5em]{\rotatebox{90}{\DAsetting{CS}{Night}}} & \rowqual{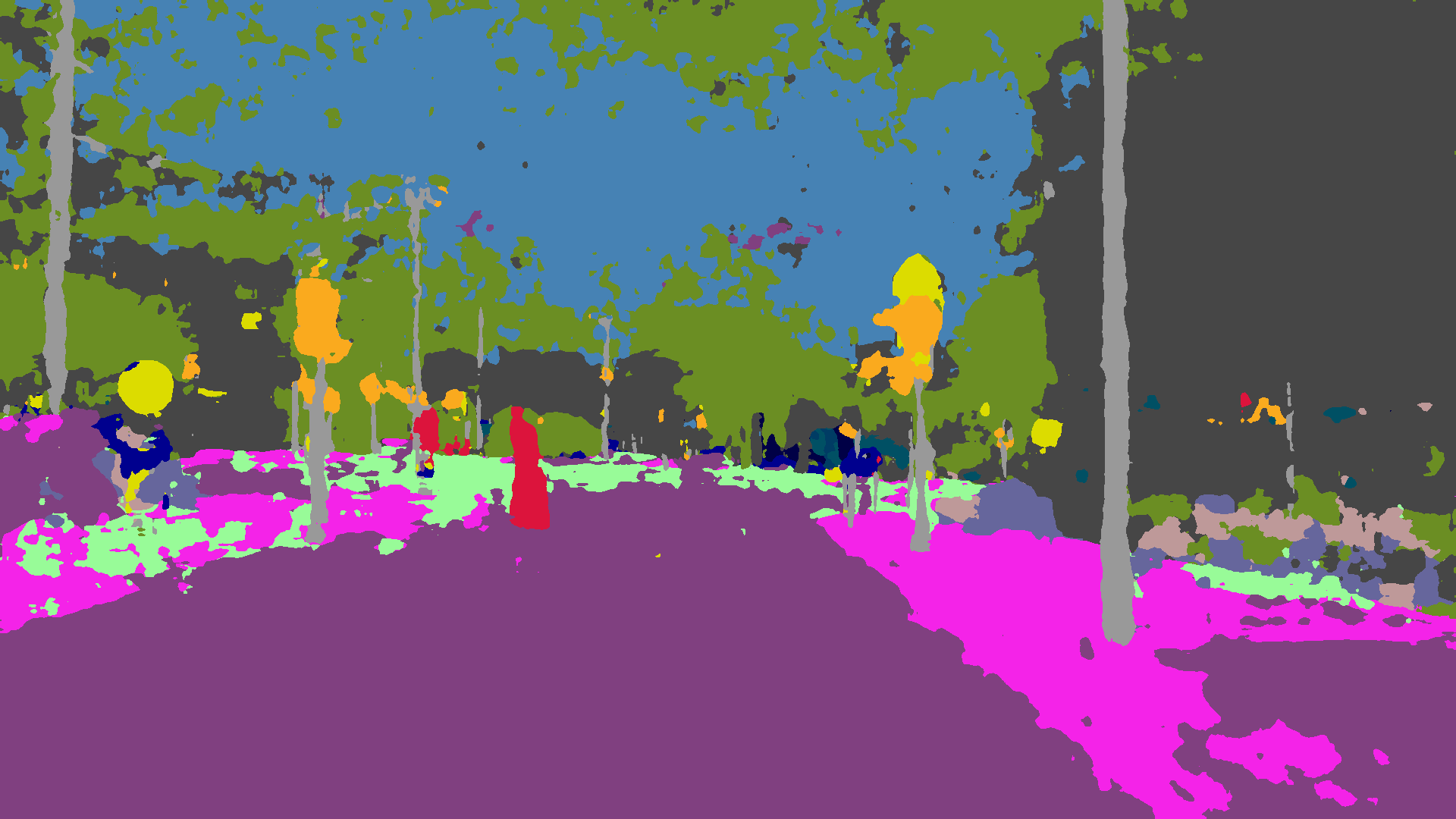}\\
        \multirow{1}{*}[4.5em]{\rotatebox{90}{\DAsetting{CS}{Rain}}} & \rowqual{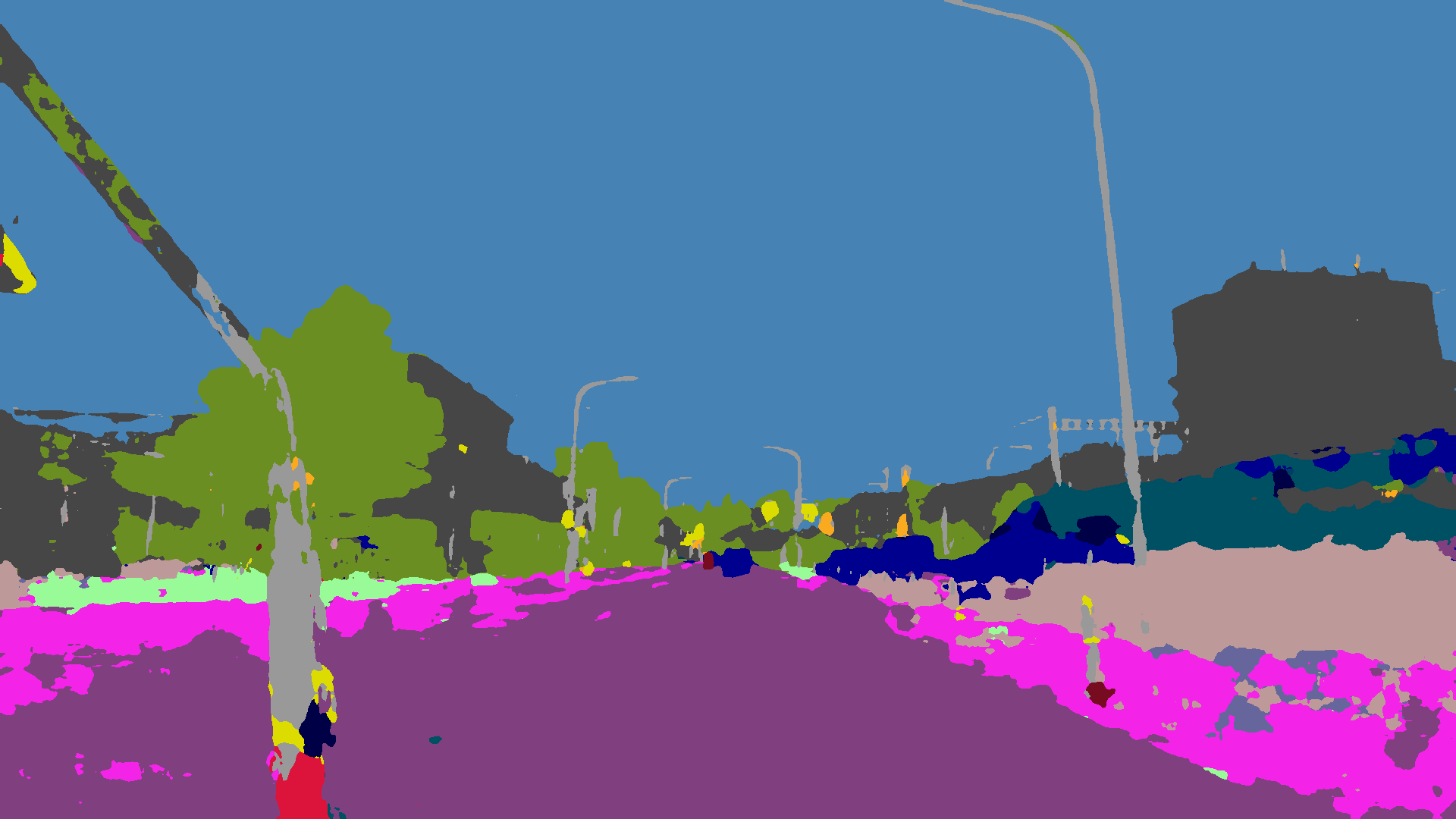}\\
        \rotatebox{90}{\DAsetting{GTA5}{Real}} & \rowqual{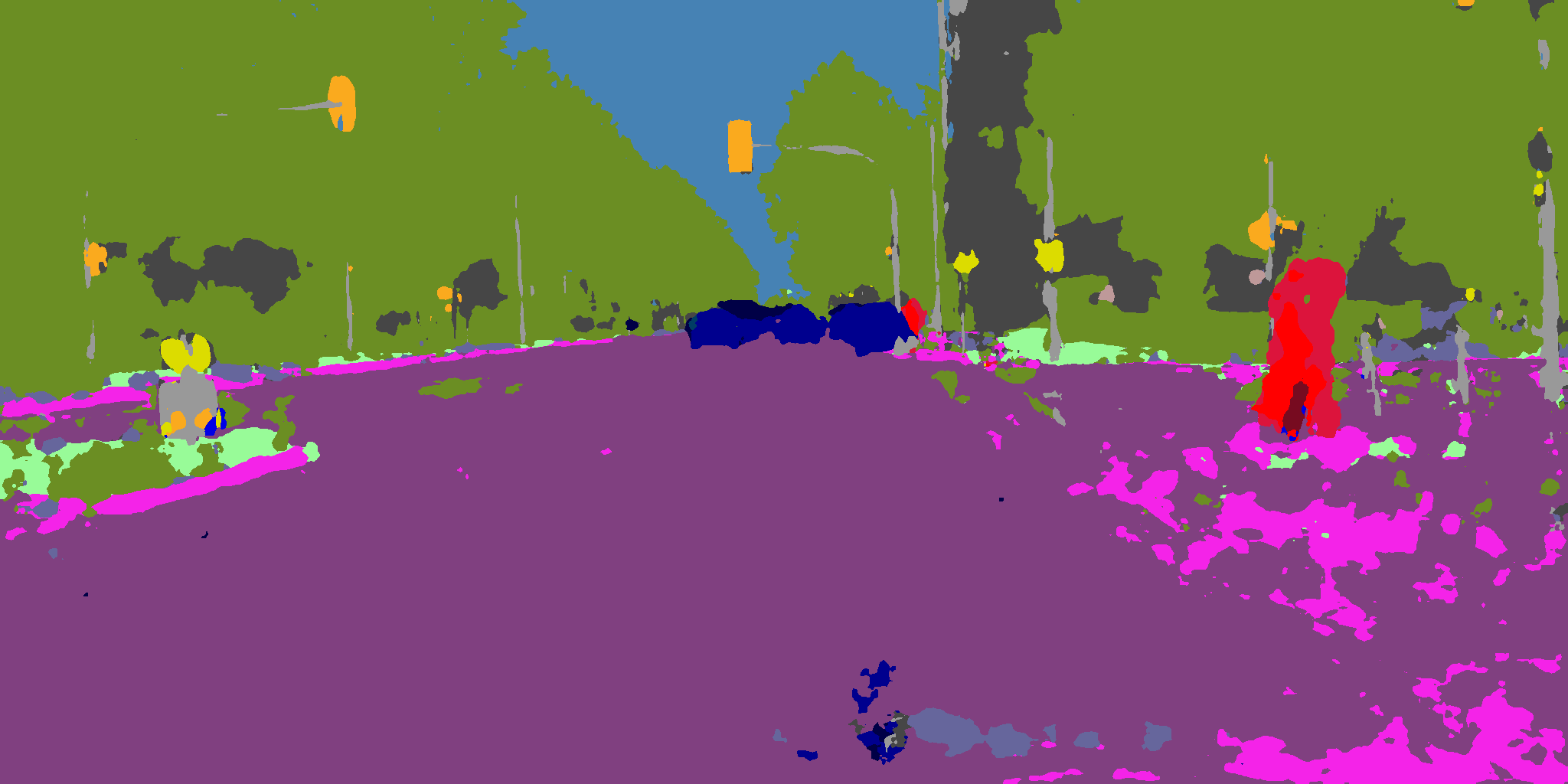}%
    \end{tabular}
    }
    \smallskip
    \caption{
    \small\textbf{Qualitative results.} (\textit{Columns 1-2}) Input images and their corresponding ground-truth segmentation; (\textit{Columns 3-4}) Segmentation predictions from the source-only model and the proposed \method model, respectively.
    }
    \label{fig:qualres}
\end{figure}

\begin{table}[t]
    \centering
  \resizebox{1.\linewidth}{!}{
  \begin{tabular}{clcc}
    \toprule
    Source & Target eval. & 1-shot SM-PPM & 0-shot \method \\ 
    \midrule
    \multirow{3}{*}{CS}
    & ACDC Night & 13.07\,/\,14.60~{\footnotesize($\Delta$=1.53)} & 18.31\,/\,\textbf{25.03} ~{\footnotesize({$\Delta$=6.72})} \\
    & ACDC Snow & 32.60\,/\,35.61 ~{\footnotesize({$\Delta$=3.01})} & 39.28\,/\,\textbf{43.90} ~{\footnotesize($\Delta$=4.62)} \\ 
    & ACDC Rain & 29.78\,/\,32.23 ~{\footnotesize($\Delta$=2.45)} & 38.20\,/\,\textbf{42.31} ~{\footnotesize($\Delta$=4.11)} \\ 
    \midrule
    GTA5 & CS & 30.48\,/\,39.32 ~{\footnotesize($\Delta$=8.84)} & 36.38\,/\,\textbf{40.08} ~{\footnotesize($\Delta$=3.70)} \\ 
    \bottomrule
  \end{tabular}}
  \smallskip\caption{
    \small\textbf{Comparison with one-shot UDA.} We compare \method to SM-PPM \citep{wu2022style}, a one-shot unsupervised domain adaptation method. The table reports semantic segmentation performance in terms of mIoU (\%) for both source-only\,/\,adapted models, along with the improvement from adaptation ($\Delta$ mIoU). For adaptation, SM-PPM (using DeepLabv2 with a ResNet-101 backbone) leverages a single target image, while PØDA (using DeepLabv3+ with a ResNet-50 backbone) relies on a target prompt and a text encoder.
    } 
  \label{tab:comparison_with_OSUDA}
\end{table}

\smallskip\noindent\textbf{Comparison to one-shot UDA (OSUDA). }
{We also compare \method~against SM-PPM \citep{wu2022style},\footnote{We use official code \url{https://github.com/W-zx-Y/SM-PPM}} a state-of-the-art OSUDA method, 
see \cref{tab:comparison_with_OSUDA}. 
The OSUDA setting allows the access to a single unlabeled target domain image for domain adaptation. In SM-PPM, this image is considered as an anchor point for target style mining.
Using $5$ randomly  selected target images, we trained, with each one, five models with different random seeds. The reported mIoUs are averaged over the 25 resulting models.
We note that the absolute results of the two models are not directly comparable due to the differences in backbone~(ResNet-101 in SM-PPM \vs ResNet-50 in \method) and in segmentation head~(DeepLabv2 in SM-PPM \vs DeepLabv3+ in \method).
We thus analyze the improvement of each method over the corresponding naive source-only baseline while taking into account the source-only performance.
We first notice that our source-only~(CLIP ResNet) performs better than \makebox{SM-PPM} source-only (ImageNet pretrained ResNet), demonstrating the overall robustness of the frozen CLIP-based model. 
In \DAsetting{Cityscapes}{ACDC}, both absolute and relative improvements of \method~over source-only are greater than the ones of SM-PPM.
Overall, \method~exhibits on par or greater improvements over \makebox{SM-PPM}, despite the fact that our method is purely zero-shot.}

\smallskip\noindent\textbf{{Qualitative results on {uncommon} conditions}.} 
{Figure \ref{fig:long_tail_quali} shows some qualitative results, training on Cityscapes, and adapting to {uncommon} conditions {never found in datasets because they are either rare (\textit{sandstorm}), dangerous (\textit{fire}), or not labeled (\textit{old movie}). For all, \method{} improves over source-only,
which demonstrates its true benefit.}}

\begin{figure}[t!]
    \centering
    \resizebox{1.0\linewidth}{!}{%
        \setlength{\tabcolsep}{0.0015\linewidth}
        \small
        \begin{tabular}{ccc}
            \textbf{Input} & \textbf{Source-only} & \textbf{\method{}}\\
            \multicolumn{3}{c}{\cellcolor{gray!34}$\prompt$ = ``driving through fire''}\\
            \includegraphics[width=0.33\linewidth]{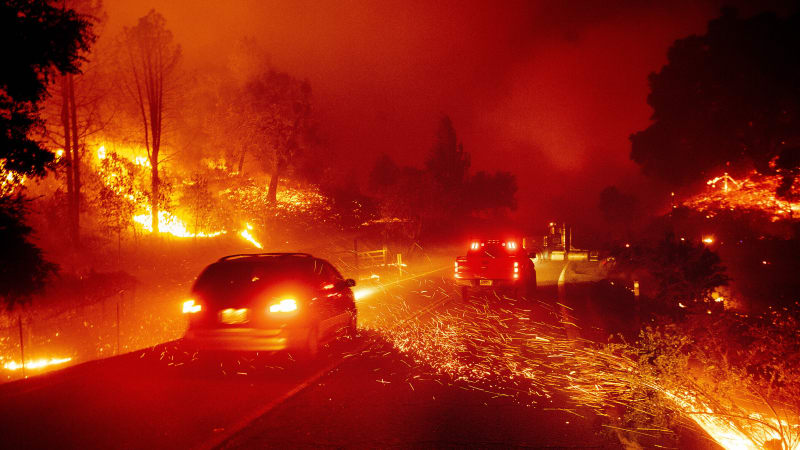} & \includegraphics[width=0.33\linewidth]{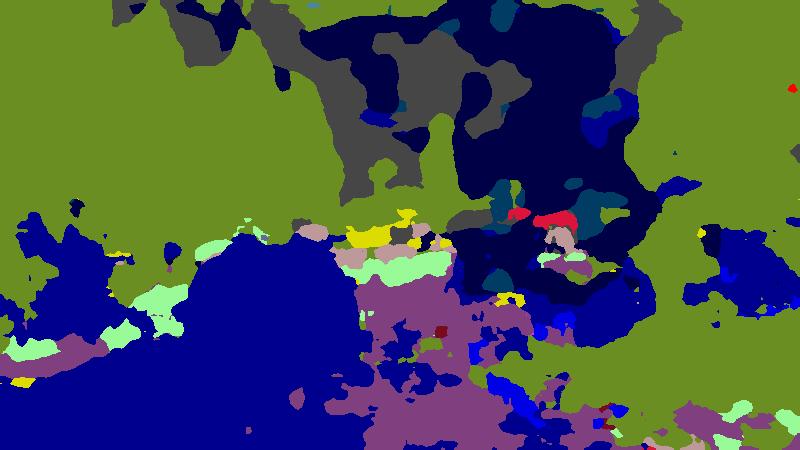} & \includegraphics[width=0.33\linewidth]{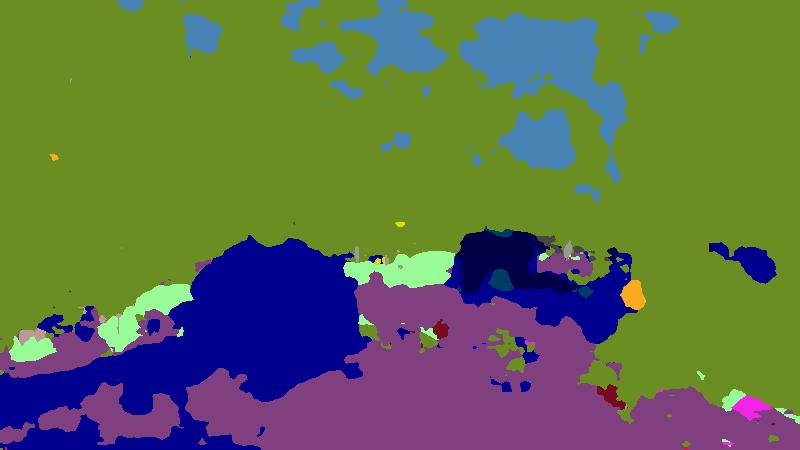}\\
            \includegraphics[width=0.33\linewidth,height=0.15\linewidth]{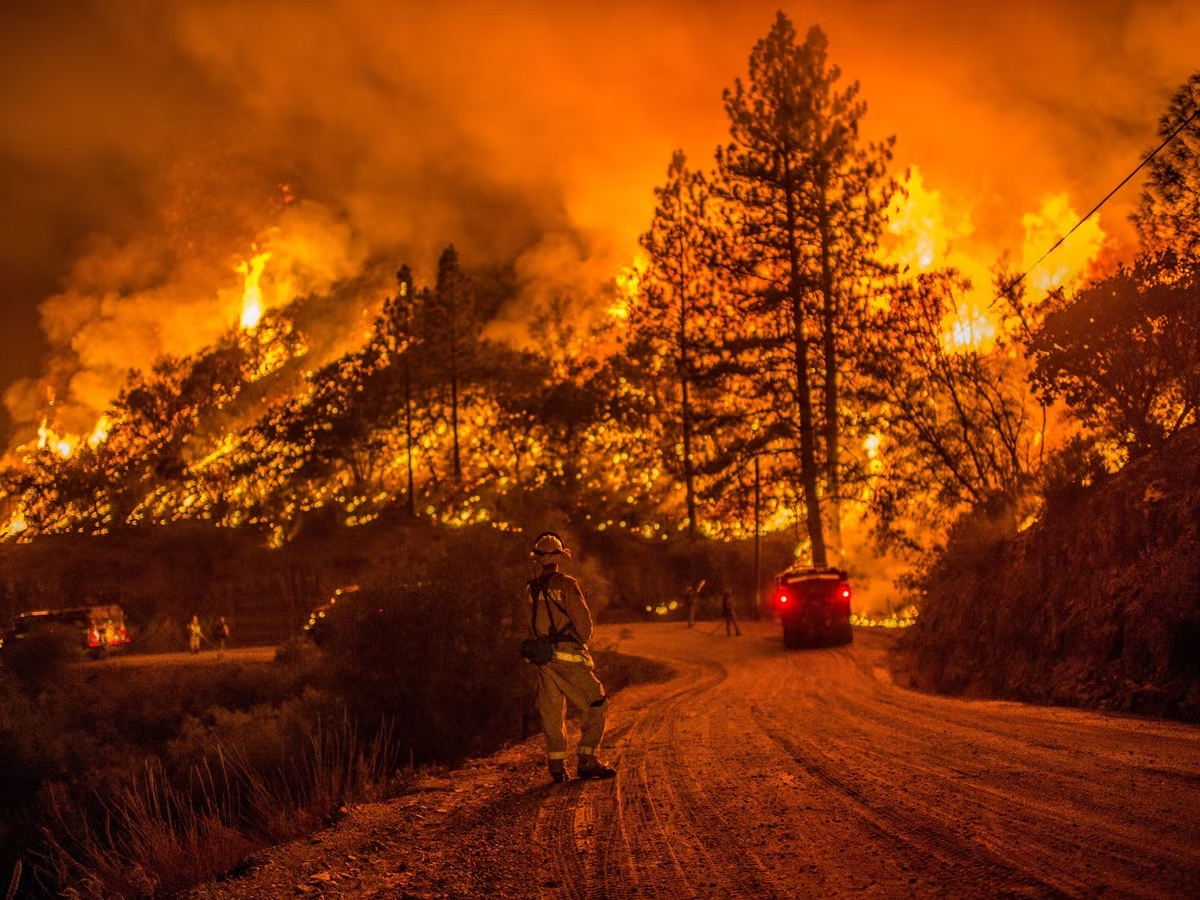} & \includegraphics[width=0.33\linewidth,height=0.15\linewidth]{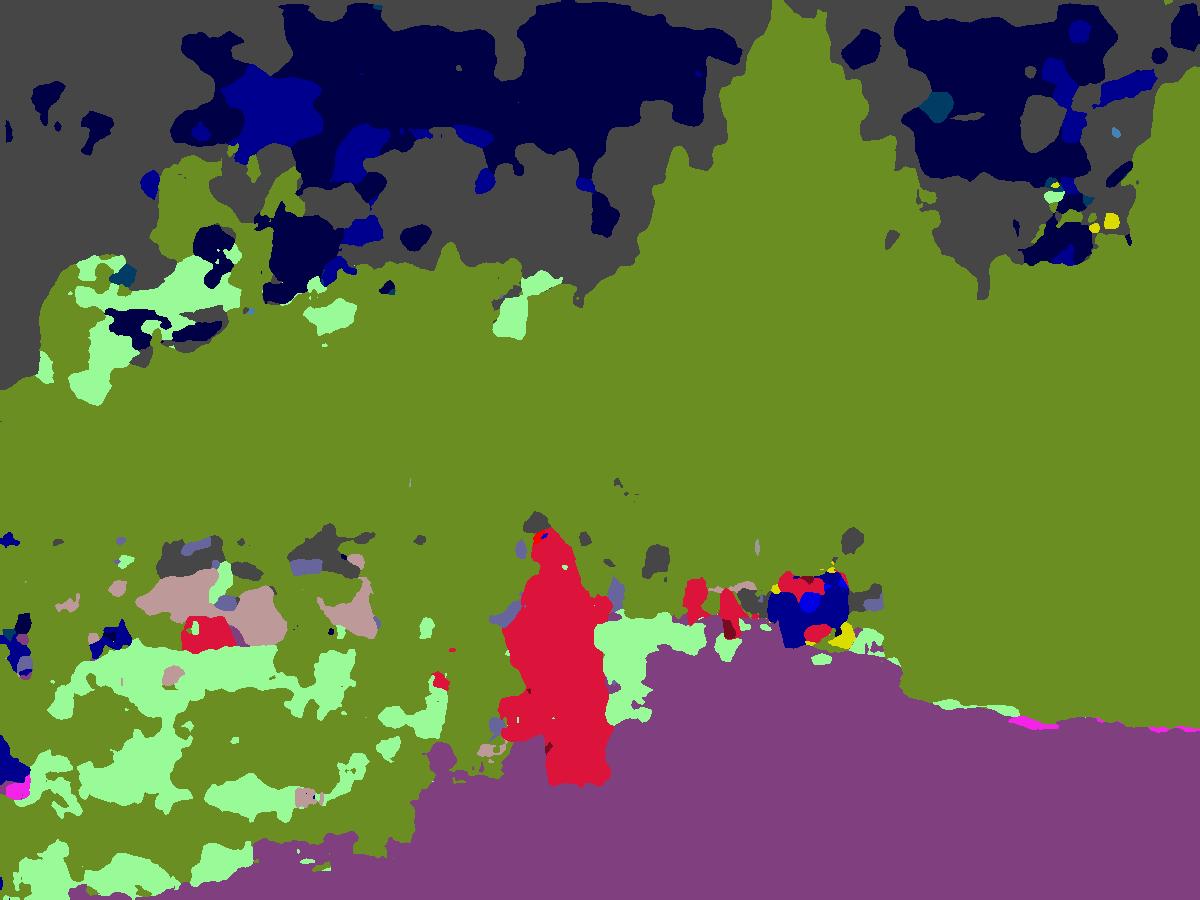} & \includegraphics[width=0.33\linewidth,height=0.15\linewidth]{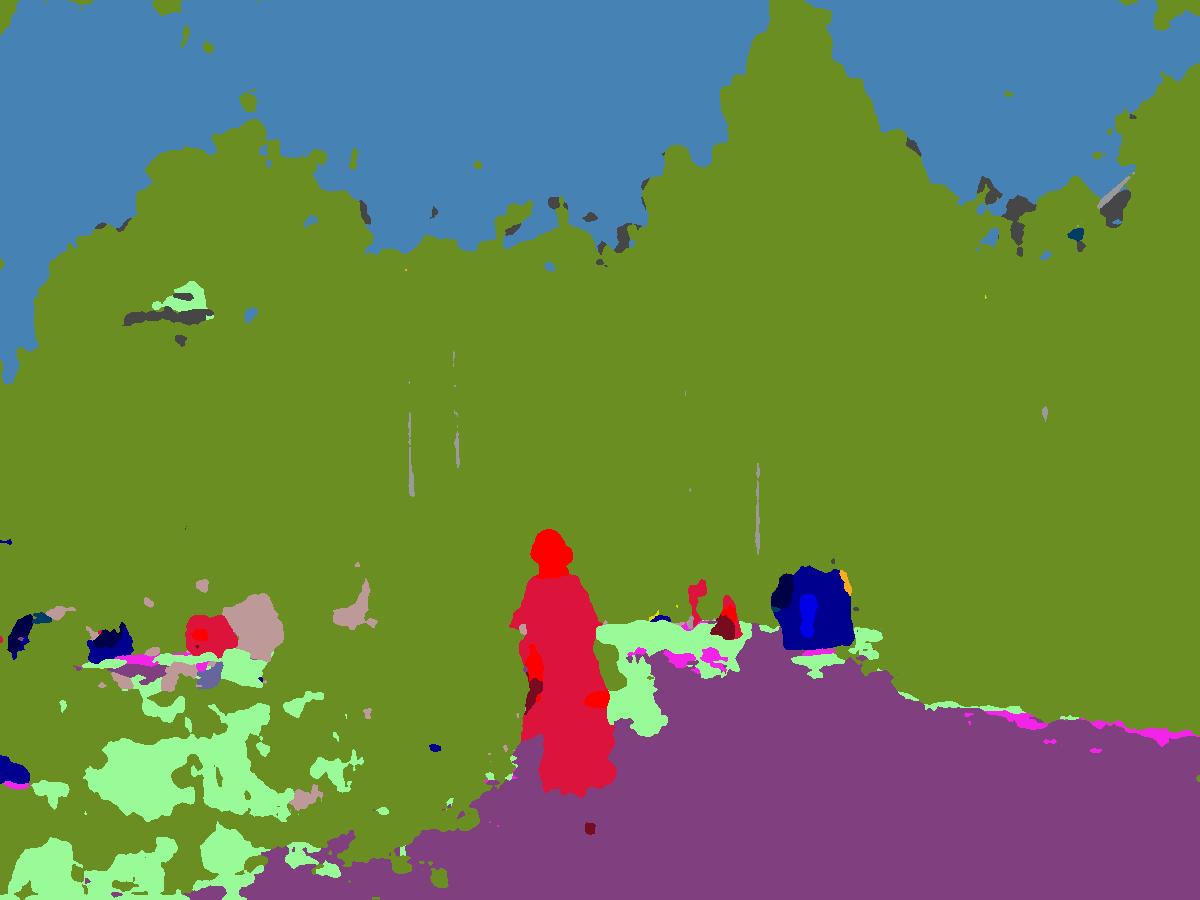}\\
            \multicolumn{3}{c}{~}\\
            \multicolumn{3}{c}{\cellcolor{gray!34}$\prompt$ = ``driving in sandstorm''}\\
            \includegraphics[width=0.33\linewidth,height=0.15\linewidth]{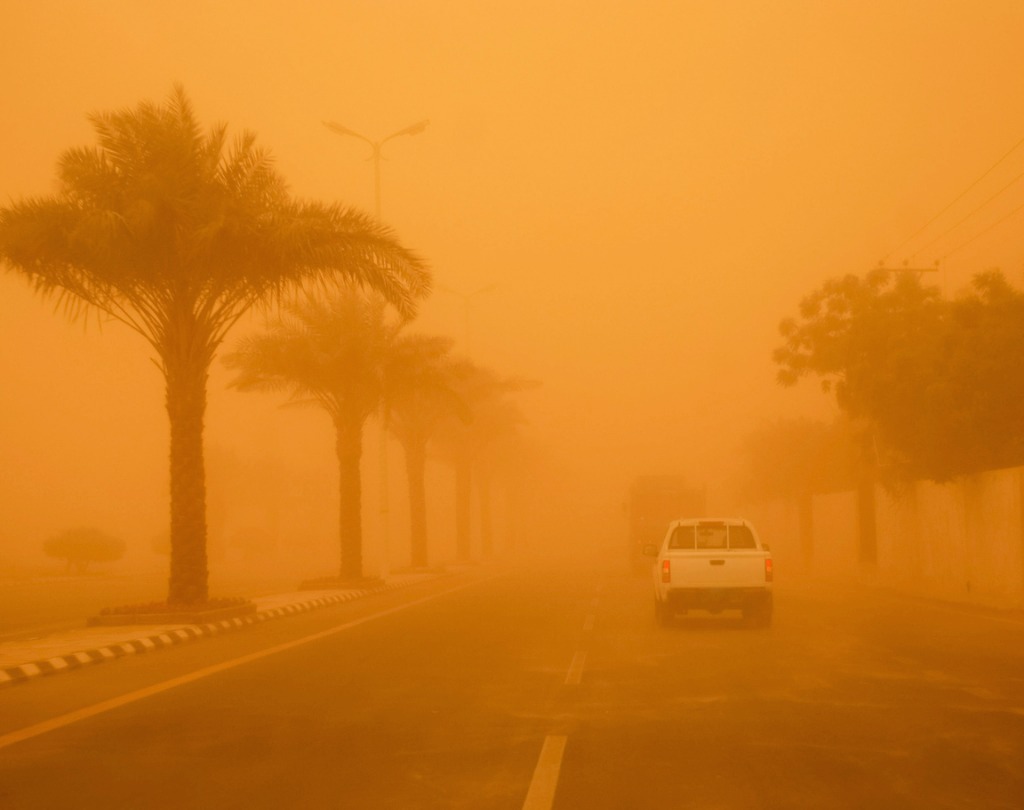} & \includegraphics[width=0.33\linewidth,height=0.15\linewidth]{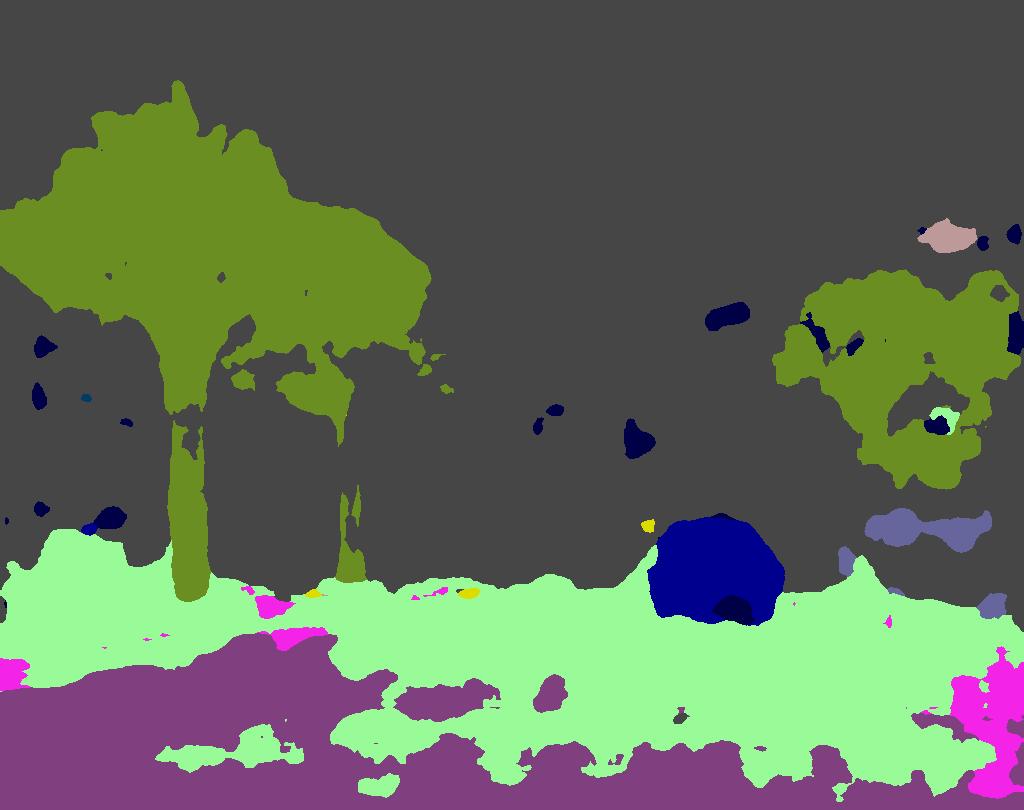} & \includegraphics[width=0.33\linewidth,height=0.15\linewidth]{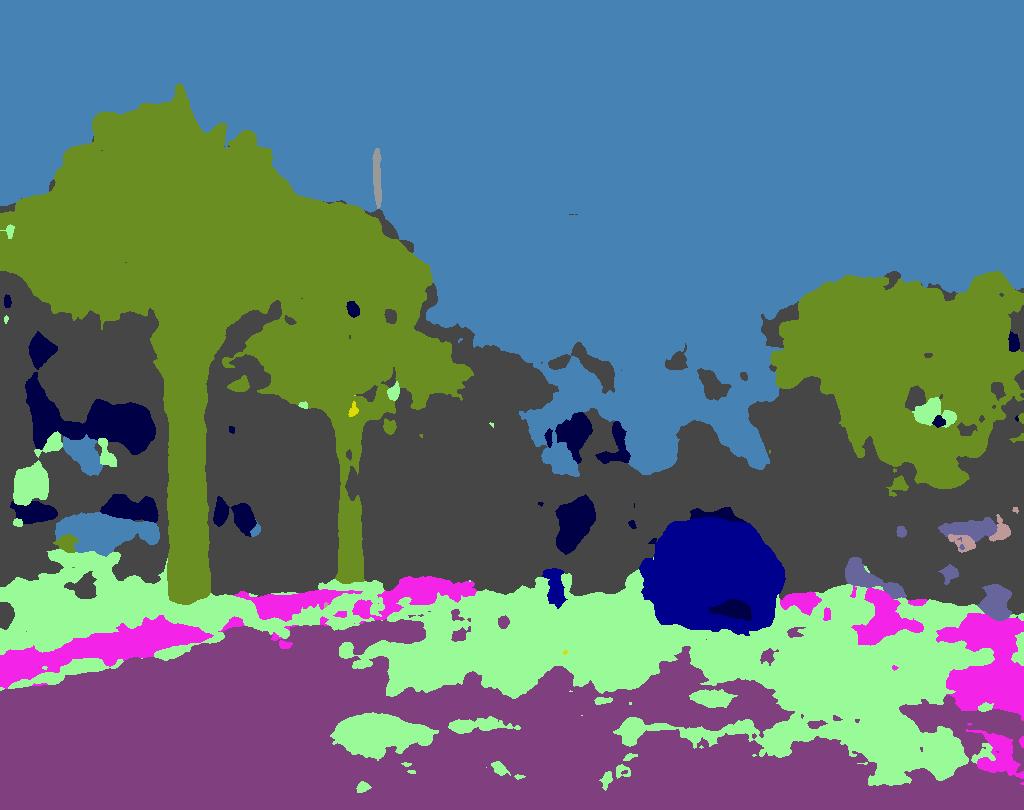}\\
            \includegraphics[width=0.33\linewidth,height=0.15\linewidth]{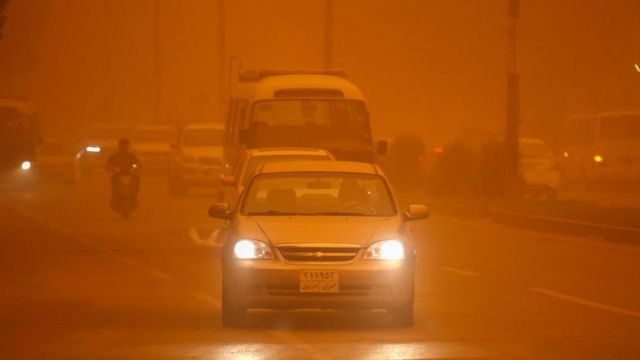} & \includegraphics[width=0.33\linewidth,height=0.15\linewidth]{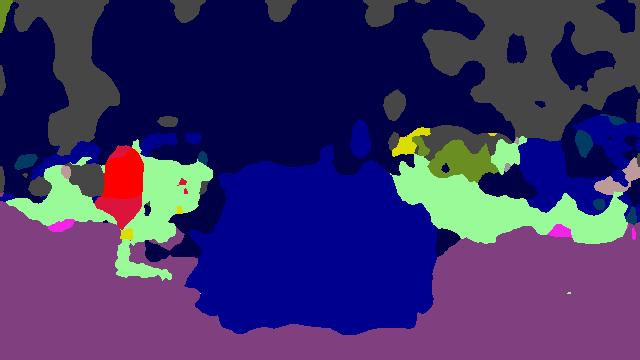} & \includegraphics[width=0.33\linewidth,height=0.15\linewidth]{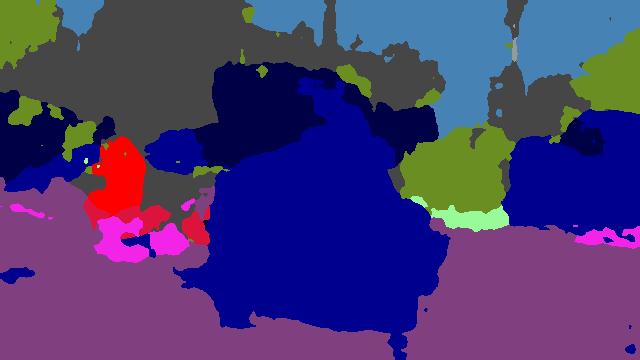}\\
            \multicolumn{3}{c}{~}\\
            \multicolumn{3}{c}{\cellcolor{gray!34}$\prompt$ = ``driving in old movie''}\\
            \includegraphics[width=0.33\linewidth,height=0.15\linewidth]{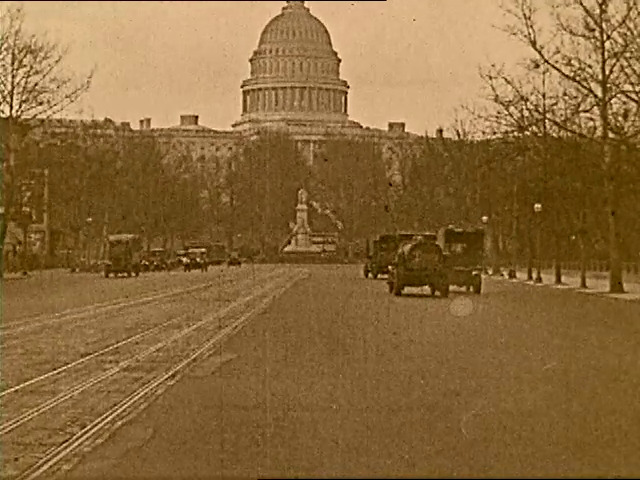} & \includegraphics[width=0.33\linewidth,height=0.15\linewidth]{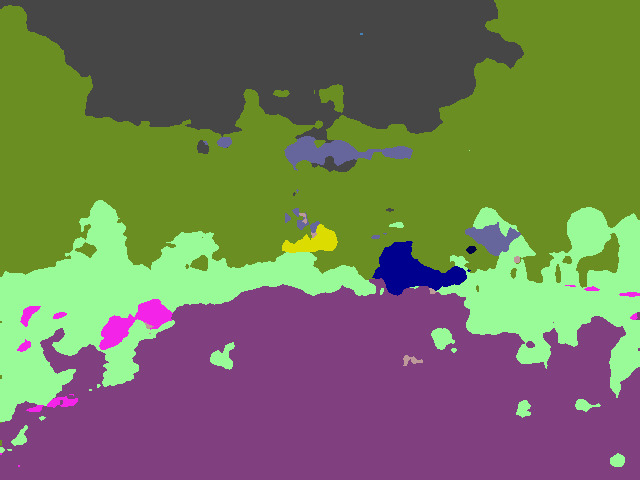} & \includegraphics[width=0.33\linewidth,height=0.15\linewidth]{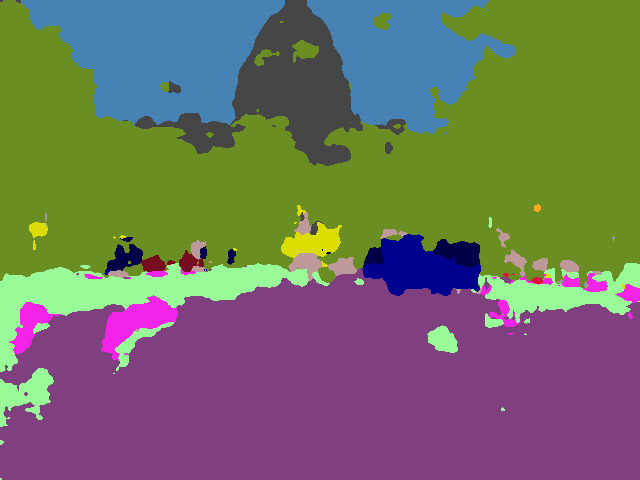}\\
            \includegraphics[width=0.33\linewidth,height=0.15\linewidth]{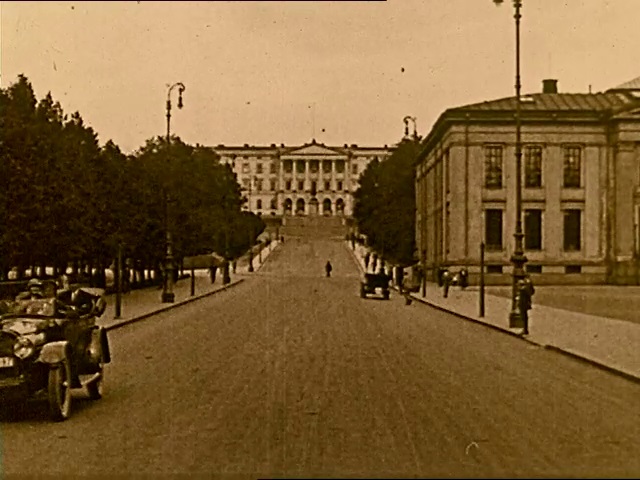} & \includegraphics[width=0.33\linewidth,height=0.15\linewidth]{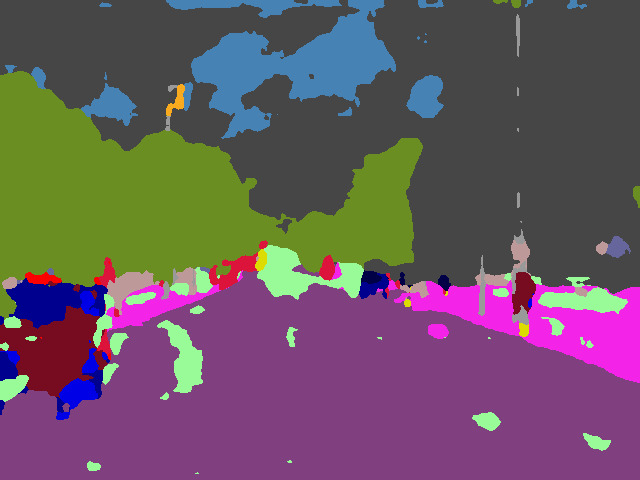} & \includegraphics[width=0.33\linewidth,height=0.15\linewidth]{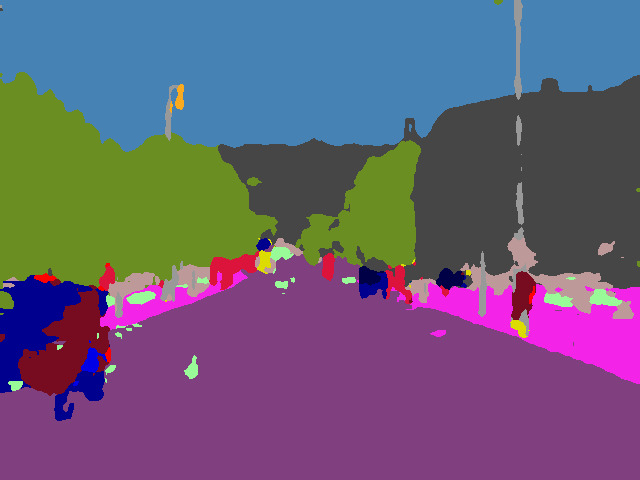}\\
        \end{tabular}
        \label{fig:long_tail}%
    }
    \smallskip\caption{
    \small\textbf{Qualitative results on uncommon conditions.}
    The source-only model is trained on Cityscapes and \method{} uses the single prompt {$\prompt$} on top of the source-only model.
    }
    \label{fig:long_tail_quali}
\end{figure}

\subsubsection{Ablation studies}
\label{sec:abla}

\noindent\textbf{Effect of prompts on \method{}.}
{Using any meaningful descriptions of the target domain as prompt $\prompt$, one should obtain similar adaptation gain with~\method{}.
{To verify this, we generate other \relprompt{}s by querying ChatGPT\footnote{OpenAI's chatbot \url{https://chatgpt.com/}} 
with \query{Give me 5 prompts that have the same exact meaning as [PROMPT]} using same prompts as in \cref{tab:main_results}.
Results in~\cref{tab:text_effect_RN50_relevant} show that 
adaptation gains are rather independent of the textual expression.
Inversely, we query \irrprompt{}s with \query{Give me 6 random prompts of length from 3 to 6 words describing a random photo}, which could result in negative transfer~(See~\cref{tab:text_effect_RN50_irrelevant}).}  
In some cases, small gains could occur; however we conjecture that such gains may originate from generalization by randomization rather than adaptation.
The generated \relprompt{}s cover a broad spectrum of conceptually synonymous expressions. For instance, the term ``driving'' may be replaced with phrases such as ``operating a vehicle'', ``navigating the roads'', ``traveling by car''. Similarly, stylistic variations like ``at night'' are reformulated as ``after sunset'', ``in darkness'', or ``in low-light conditions''. As shown in \cref{tab:text_effect_RN50_relevant},  these alternative phrasings maintain the effectiveness of \method{}, provided that they convey the same underlying semantic meaning. In contrast, \irrprompt{}s—those with semantically unrelated content—demonstrate noticeably reduced performance.}

\begin{table}
\setlength{\tabcolsep}{0.01\linewidth}
\renewcommand{\arraystretch}{1.2}
\centering
\newcolumntype{H}{>{\setbox0=\hbox\bgroup}c<{\egroup}@{}}
\newcommand{\MPrompt}[1]{\cellcolor{gray!34}``#1''}
\newcommand{\GPTRel}[1]{\multicolumn{1}{>{\columncolor{gray!34}}c}{\color{colrelprompt!100}\tabularCenterstack{c}{``#1''}}}
\newcommand{\GPTIrr}[1]{\multicolumn{4}{>{\columncolor{gray!34}}c}{\color{colirrprompt!100}\tabularCenterstack{c}{``#1''}}}
\newcommand{\best}[1]{\textbf{#1}}
\resizebox{1.0\linewidth}{!}{
    \begin{tabular}{ccccc} \toprule
	Method & ACDC Night & ACDC Snow & ACDC Rain & GTA5\quad\quad\quad{}\\ \midrule 
	\makecell{Source\\only} & 18.31 & 39.28 & 38.20 & 39.59\\ \midrule
		\makecell{$\prompt$} & \MPrompt{driving at night} & \MPrompt{driving in snow} & \MPrompt{driving under rain} & \MPrompt{driving in a game}\\
		& 25.03\vartn{0.48}  & 43.90\vartn{0.53} & 42.31\vartn{0.55} & 41.07\vartn{0.48}\\ \midrule
		\multirow{26}{*}{\rotatebox{90}{\hspace{-21.5em} {\color{colirrprompt!100} \Large Irrelevant}\quad\fbox{ChatGPT-generated}\quad {\color{colrelprompt!100} \Large Relevant}}} & 
            \GPTRel{operating a\\vehicle\\after sunset} & \GPTRel{operating a\\vehicle in\\snowy conditions} & \GPTRel{operating a\\vehicle in\\ wet conditions} & \GPTRel{piloting a\\vehicle in\\a virtual world}\\
		& 24.38\vartn{0.37} & 44.33\vartn{0.36} & 42.21\vartn{0.47} &  41.25\vartn{0.40}\\
		& \GPTRel{driving during\\the nighttime\\hours} & \GPTRel{driving on\\snow-covered\\roads} & \GPTRel{driving on\\rain-soaked\\roads} & \GPTRel{controlling a car\\ in a digital\\simulation} \\
		& \textbf{25.22}\vartn{0.64} & 43.56\vartn{0.62} & \textbf{42.51}\vartn{0.33} & 41.19\vartn{0.14} \\
		& \GPTRel{navigating\\the roads\\in darkness} & \GPTRel{piloting a\\vehicle in\\snowy terrain} & \GPTRel{navigating\\through rainfall\\while driving} & \GPTRel{maneuvering\\a vehicle in a\\ computerized racing\\ experience'} \\
		& 24.73\vartn{0.47}  & \textbf{44.67}\vartn{0.18} & 41.11\vartn{0.69} & 40.34\vartn{0.49}\\
		& \GPTRel{driving in\\low-light\\conditions} & \GPTRel{driving in\\wintry\\precipitation} & \GPTRel{driving in\\inclement\\weather} & \GPTRel{operating\\a transport \\in a video game\\environment} \\
		& 24.68\vartn{0.34}  & 43.11\vartn{0.56} & 40.68\vartn{0.37} & 41.34\vartn{0.42} \\
		& \GPTRel{travelling\\by car\\after dusk} & \GPTRel{travelling\\by car in\\a snowstorm} & \GPTRel{travelling by\\car during\\a downpour} & \GPTRel{navigating a\\machine through\\a digital\\driving simulation} \\
		& 24.89\vartn{0.24} & 43.83\vartn{0.17} & 42.05\vartn{0.35} & \textbf{41.86}\vartn{0.10} \\ \cmidrule(lr){2-2} \cmidrule(lr){3-3} \cmidrule(lr){4-4}
\cmidrule(lr){5-5}
		& \textit{24.82} & \textit{43.90} & \textit{41.81} & \textit{41.18} \\ \cmidrule{2-5}
            & \GPTIrr{mesmerizing northern lights display}\\
		&  20.05\vartn{0.77} &  40.07\vartn{0.66} & 38.43\vartn{0.82} & 37.98\vartn{0.31} \\
		& \GPTIrr{playful dolphins in the ocean}\\
		& 20.11\vartn{0.31} & 39.87\vartn{0.26} & 38.56\vartn{0.58} & 37.05\vartn{0.31}\\
		& \GPTIrr{breathtaking view from mountaintop}\\
		& 20.65\vartn{0.33} & 42.08\vartn{0.28} & 40.05\vartn{0.52} & 40.09\vartn{0.23}\\
		& \GPTIrr{cheerful sunflower field in bloom}\\
		& 21.10\vartn{0.50} & 39.85\vartn{0.68} & 40.09\vartn{0.41} & 37.93\vartn{0.55}\\
		& \GPTIrr{dramatic cliff overlooking the ocean}\\
		& 20.09\vartn{0.98} & 38.20\vartn{0.54} & 38.48\vartn{0.37} & 37.57\vartn{0.46}\\
		& \GPTIrr{majestic eagle in flight over mountains}\\
 		& 20.70\vartn{0.38} & 39.60\vartn{0.27} & 40.38\vartn{0.86} & 38.52\vartn{0.21}\\ 
            \cmidrule(lr){2-2} \cmidrule(lr){3-3} \cmidrule(lr){4-4}
\cmidrule(lr){5-5}
            & \textit{20.45} & \textit{39.95} & \textit{39.33} & \textit{38.19} \\ \bottomrule
    \end{tabular}}
    \smallskip\caption{
    \small\textbf{Effect of prompts on \method{}.}
    This table presents results for three types of prompts: our hand-crafted prompt $\prompt{}$ (\textit{top}), \setlength{\fboxsep}{1pt}\colorbox{gray!34}{\relprompt{}} generated by ChatGPT (\textit{middle}), and \setlength{\fboxsep}{1pt}\colorbox{gray!34}{\irrprompt{}} also generated by ChatGPT (\textit{bottom}). For more details, refer to \cref{sec:abla}. The best results (in \textbf{bold}) are consistently achieved with {\color{colrelprompt!100}relevant prompts}, which also yield higher average mIoU scores (\textit{italic}).
    }
    \label{tab:text_effect_RN50_relevant}
    \label{tab:text_effect_RN50_irrelevant}
\end{table}

\noindent\textbf{Effect of prompt length on \method{}.} 
\noindent In addition to semantic consistency, we investigate the effect of prompt length by generating a new set of \relprompt{}s querying ChatGPT with: \query{Give me 5 prompts that have the same exact meaning as [PROMPT] with increasing length.} Results presented in \cref{tab:prompt_length} indicate consistent improvements over the source-only model across all prompt lengths, with no clear correlation between length and performance. Notably, some of the longest prompts (last row in \cref{tab:prompt_length}) yield slightly better results, suggesting that additional descriptive detail can occasionally enhance the adaptation effect, though not uniformly.

\begin{table}
\setlength{\tabcolsep}{0.01\linewidth}
\renewcommand{\arraystretch}{1.2}
\centering
\newcolumntype{H}{>{\setbox0=\hbox\bgroup}c<{\egroup}@{}}
\newcommand{\MPrompt}[1]{\cellcolor{gray!34}``#1''}
\newcommand{\GPTRel}[1]{\multicolumn{1}{>{\columncolor{gray!34}}c}{\color{colrelprompt!100}\tabularCenterstack{c}{``#1''}}}
\newcommand{\GPTRelrev}[1]{\multicolumn{1}{>{\columncolor{gray!34}}c}{\color{black}\tabularCenterstack{c}{``#1''}}}
\newcommand{\GPTIrr}[1]{\multicolumn{4}{>{\columncolor{gray!34}}c}{\color{colirrprompt!100}\tabularCenterstack{c}{``#1''}}}
\newcommand{\best}[1]{\textbf{#1}}
\resizebox{1.0\linewidth}{!}{
    \color{black}
    \begin{tabular}{ccccc} \toprule
	Method & ACDC Night & ACDC Snow & ACDC Rain & GTA5\quad\quad\quad{}\\ \midrule 
	\makecell{Source\\only} & 18.31 & 39.28 & 38.20 & 39.59\\ \midrule
    \multirow{10}{*}{\rotatebox{90}{\hspace{-17em}{\large $\longleftarrow$ Longer}\quad\fbox{ChatGPT-generated}}} \\
		\makecell{$\prompt$} & \MPrompt{driving at night} & \MPrompt{driving in snow} & \MPrompt{driving under rain} & \MPrompt{driving in a game}\\
		& 25.03\vartn{0.48}  & 43.90\vartn{0.53} & 42.31\vartn{0.55} & 41.07\vartn{0.48}\\ \midrule
        & \GPTRelrev{night drive} & \GPTRelrev{snowy drive} & \GPTRelrev{rainy drive} & \GPTRelrev{game driving}\\
		& 24.97\vartn{0.52} & 42.61\vartn{0.53} & 41.26\vartn{0.52} & 41.03\vartn{0.40} \\
		& \GPTRelrev{driving through \\ the night} & \GPTRelrev{driving \\ through snow} & \GPTRelrev{driving through rain} & \GPTRelrev{driving in \\ a video game} \\
		&  24.71\vartn{0.61} & 42.92\vartn{0.28} & \best{42.43}\vartn{0.37} & 40.85\vartn{0.45} \\
		& \GPTRelrev{operating a \\ vehicle after \\ dark} & \GPTRelrev{navigating a \\ vehicle in \\ snowy conditions} & \GPTRelrev{navigating a \\ vehicle in \\ rainy weather} & \GPTRelrev{controlling a \\ car in \\ a virtual game} \\
		& 25.63\vartn{0.75} & 43.52\vartn{0.26} & 42.19\vartn{0.45} & 40.67\vartn{0.42} \\
		& \GPTRelrev{cruising down \\ the road \\ in the nighttime} & \GPTRelrev{carefully driving \\ a car \\ while snow \\ is falling} & \GPTRelrev{driving a \\ car while \\ raindrops fall \\ on the windshield} & \GPTRelrev{simulating vehicle \\ driving within \\ a video game} \\
		&  24.87\vartn{0.38} & 43.33\vartn{0.52} & 41.56\vartn{0.50} & 41.07\vartn{0.51} \\
		& \GPTRelrev{traveling by \\ car under \\ the stars \\ during the \\ late hours \\ of the night} & \GPTRelrev{Operating a \\ vehicle on \\ snow-covered \\ roads during \\ a winter storm} & \GPTRelrev{traveling \\ by car \\ on wet roads \\ as steady \\ rain pours \\ from the sky} & \GPTRelrev{engaging in \\ interactive car \\ driving as \\ part of a \\ gaming experience} \\
		& \best{25.82}\vartn{0.43} & \best{44.38}\vartn{0.43} & 42.32\vartn{0.36} & \best{41.72}\vartn{0.40}\\

      \bottomrule
    \end{tabular}}
    \smallskip\caption{
    \small\textbf{Effect of prompt length on \method{}.} This table presents results for prompts with increasing length (from top to bottom), generated by ChatGPT. For more details, refer to \cref{sec:abla}.
    }
    \label{tab:prompt_length}
\end{table}

\smallskip\noindent\textbf{Impact of selected layers for feature augmentation.}~DeepLabV3+ 
segmenter takes as inputs both low-level features from \textit{Layer1} and high-level features from \textit{Layer4}.
In~\method, we only augment the \textit{Layer1} features (by optimizing PIN) and forward them through remaining layers 2-4 to obtain the \textit{Layer4} features. 
The input to the segmentation head is the concatenation of both.
We study in \cref{tab:layer_augment} if one should augment other features in addition to the ones in \textit{Layer1}: 
we observe 
the best performance with only \textit{Layer1} augmentation.
We conjecture that it is important to preserve the consistency between the two inputs to the classifier, \ie, \textit{Layer4} features should be derived from the augmented ones of \textit{Layer1}.

\noindent We highlight that augmenting low-level features is detrimental for style manipulation. A decade of research in neural style transfer with CNNs has solidified the principle that CNN low-level features capture localized structures and visual style, whereas deeper features encodes global semantics. 
Methodologically, style manipulation is typically achieved through two avenues: calculating feature correlations via Gram matrices~\citep{gatys2015neural,gatys2016image}, or manipulating feature channel statistics~\citep{huang2017arbitrary,pan2018two,fan2023towards}. Following the latter, we extract stylistic information from textual prompts by optimizing low-level channel statistics. This preserves the content features, avoiding the degradation CLIP's semantic representations suggested by the lower performance in~\cref{tab:layer_augment} when PIN is applied in deeper blocks. 
\begin{table}[t]
  \setlength{\tabcolsep}{0.025\linewidth}
  \centering
  \begin{tabular}{ccccc}
    \toprule
    \textit{Layer1} & \textit{Layer2} & \textit{Layer3} & \textit{Layer4} &  ACDC Night \\
    \midrule
    \checkmark & \xmark & \xmark & \xmark & \textbf{25.03}\vartn{0.48} \\
    \checkmark & \checkmark & \xmark & \xmark & 23.43\vartn{0.51} \\
    \checkmark & \xmark & \checkmark & \xmark & 22.93\vartn{0.53} \\
    \checkmark & \xmark & \xmark & \checkmark & 21.05\vartn{0.55} \\
    \bottomrule
  \end{tabular}
   \smallskip\caption{
   \small\textbf{Impact of selected layers for feature augmentation.} This table shows the semantic segmentation performance (mIoU) of \method{} for \DAsetting{day}{night} adaptation, using different ResNet layers for feature augmentation. In addition to augmenting features of \textit{Layer1} ({Row\,1)}, one can augment \textit{Layer2}, \textit{Layer3}, or \textit{Layer4} features ({Rows\,2-4}).
   }
 \label{tab:layer_augment}
\end{table}

\smallskip\noindent\textbf{Number of mined styles.~}
{In our experiments, $|\stot{\mc{S}}| = |\src{\mc{D}}|$. Here we study the effect of changing the number $|\stot{\mc{S}}|$ of styles on the target domain performance. By performing ablation on \DAsetting{CS}{Night} with $|\stot{\mc{S}}|=$ 1, 10, 100, 1000, 2975 (i.e., $|\src{\mc{D}}|$), we obtain 16.00\vartn{5.01}, 22.04\vartn{1.24}, 23.90\vartn{0.96}, 24.27\vartn{0.70}, 25.03\vartn{0.48} (mIoU \%) respectively. 
{For $|\stot{\mc{S}}| < |\src{\mc{D}}|$, the styles are sampled randomly from $\src{\mc{D}}$ and results are reported in average on 5 different samplings. Interestingly, we observe that the variance decreases with the increase of $|\stot{\mc{S}}|$. Results also suggest that only few styles (\eg, $|\stot{\mc{S}}|= 10$) could be sufficient for feature translation, similarly to few-shot image-to-image translation~\citep{pizzati2022manifest}, though at the cost of higher variance.
}}

\smallskip\noindent\textbf{Partial unfreezing of the backbone.~}
While our experiments use a frozen backbone due to the observed good out-of-distribution performance~(\cref{tab:baselines}), we highlight that during training only \textit{Layer1 must} be frozen to preserve its activation space where augmentations are done; the remaining three layers could be optionally fine-tuned. Results in ~\cref{tab:p_unfrozen} show that freezing the whole backbone (\ie, \text{Layer1-4}) achieves the best results.
In all cases, \method consistently improves the performance over source-only.

\begin{table}[h]
\setlength{\tabcolsep}{0.01\linewidth}
\centering
 \resizebox{1.0\linewidth}{!}{
	\begin{tabular}{lllll}
		\toprule
		  Method & Night & Snow & Rain & GTA5\\
		\midrule
        src-only* &  18.31 & 39.28 & 38.20 & 39.59 \\
        \method{}* & \textbf{25.03}\vartn{0.48} & \textbf{43.90}\vartn{0.53} & \textbf{42.31}\vartn{0.55} & \textbf{41.07}\vartn{0.48} \\
        \midrule
        src-only (\textit{Layer1} \texttwemoji{2744}) &  9.60 & 30.99 & 30.89\textcolor{white}{\vartn{0,00}} & 29.38 \\
        \method{} (\textit{Layer1} \texttwemoji{2744}) & 19.43\vartn{0.69} & 37.80\vartn{2.65} & 40.71\vartn{1.06} & 39.09\vartn{1.23}\\
        \bottomrule
	\end{tabular}
 }
    \smallskip\caption{
    \small\textbf{Partial unfreezing of the backbone.} \method{} results when only \textit{Layer1} is frozen (\texttwemoji{2744}). In contrast, the models marked with * (reported in \cref{tab:main_results}) freeze the entire backbone,\ie, \textit{Layers 1 through 4}.
 }
    \label{tab:p_unfrozen}
\end{table}

Note that for fine-tuning the backbone, different hyperparameters, optimizer, and/or fine-tuning strategies might bring performance improvement \textit{w.r.t.} to the current results. This is beyond the scope of this work, which aims to show that PIN is a plug-and-play module that can be used for any CNN-based vision-language pretrained model. Other works have for example shown that PIN can be used for state-of-the-art language-driven domain generalization~\citep{fahes2024simple}.

We show in Appendix~\ref{sec:extensiontasks} that \method{} can be applied on object detection and image classification. Additional ablations on PIN initialization and on training for adaptation are included in Appendix~\ref{sec:additional_exp}.

\subsection{\methodconcept{}}
\label{sec:exp-poda-concept}

Our initial prompts are intuitive descriptions verbalizing the target domains, \eg, ``driving at night'', ``driving under rain''.
	Each prompt is the natural combination of a global {\textit{content}}, (\ie, ``driving'') and a target style condition (\eg, ``at night'', ``under rain'').
	Equivalently, the {\color{colrelprompt!100}relevant prompts} queried using ChatGPT (See~\cref{tab:text_effect_RN50_relevant}) follow the same construction.
	For instance, in ``navigating the roads in darkness'', the {concept} is expressed by ``navigating the roads'' and the style condition is described by ``in darkness''.
	The presence of both elements in the target prompt is crucial for mining relevant styles.
	Indeed, in the rows of \method of~\cref{tab:prompt_const}, we show that while using either element alone improves over source-only, yet better results are obtained when combining both.
	Interestingly, optimizing towards ``driving'' improves all the target performances without any use of a textual style condition.
	Such prompt representation could describe driving scene images with all the styles included, bringing some generalization effects.

\begin{table}
\setlength{\tabcolsep}{0.01\linewidth}
\renewcommand{\arraystretch}{1.2}
\centering
\newcolumntype{H}{>{\setbox0=\hbox\bgroup}c<{\egroup}@{}}
\newcommand{\MPrompt}[1]{\cellcolor{gray!34}``#1''}
\newcommand{\GPTRel}[1]{\multicolumn{1}{>{\columncolor{gray!34}}c}{\color{colrelprompt!100}\tabularCenterstack{c}{``#1''}}}
\newcommand{\GPTIrr}[1]{\multicolumn{4}{>{\columncolor{gray!34}}c}{\color{colirrprompt!100}\tabularCenterstack{c}{``#1''}}}
\newcommand{\best}[1]{\textbf{#1}}
\resizebox{1.0\linewidth}{!}{
    \begin{tabular}{ccccc}
    \toprule
    Method & ACDC Night & ACDC Snow & ACDC Rain & GTA5 \quad\quad\quad{}\\ \midrule 
	\makecell{Source\\only} & 18.31 & 39.28 & 38.20 & 39.59\\ \midrule
 \multirow{7}{*}{\hspace{2em}\method{}}& \MPrompt{driving} & \MPrompt{driving} & \MPrompt{driving} & \MPrompt{driving}\\
    & 22.81\vartn{0.34}  & 43.34\vartn{0.42} & 40.65\vartn{0.99}  & 40.33\vartn{0.46} \\\cmidrule{2-5}
& \MPrompt{night} & \MPrompt{snow} & \MPrompt{rain} & \MPrompt{game} \\
    & 23.74\vartn{0.55}  & 40.45\vartn{0.37}  & 39.74\vartn{0.35}  & 40.69\vartn{0.38}\\\cmidrule{2-5}
& \MPrompt{driving at night} & \MPrompt{driving in snow} & \MPrompt{driving under rain} & \MPrompt{driving in a game}\\\cmidrule{2-5}
    & \textbf{25.03}\vartn{0.48}  & \textbf{43.90}\vartn{0.53} & \textbf{42.31}\vartn{0.55} & \textbf{41.07}\vartn{0.48}\\ \bottomrule
    \multirow{4}{*}{\methodconcept{}} & \MPrompt{$\concept{}$} & \MPrompt{$\concept{}$} & \MPrompt{$\concept{}$} & \MPrompt{$\concept{}$}\\
    & 21.90\vartn{0.19} & {44.04}\vartn{0.34} & \textbf{43.25}\vartn{0.45} & {42.54}\vartn{0.29} \\\cmidrule{2-5}
    & \MPrompt{$\concept{}$ at night} & \MPrompt{$\concept{}$ in snow} & \MPrompt{$\concept{}$ under rain} & \MPrompt{$\concept{}$ in a game}\\
    & \textbf{25.16}\vartn{0.31}  & \textbf{44.72}\vartn{0.24} & {42.78}\vartn{0.35} & \textbf{42.90}\vartn{0.17}\\ \midrule
    
    \end{tabular}
    }
    \smallskip\caption{
    \small\textbf{Effect of prompt construction.} mIoU (\%) performance on the target datasets under different prompt configurations. For \method{}, we evaluate prompts based on the text concept ``driving'', the target style descriptions (\eg, ``night''), and their combination (prompt $\prompt$, \eg, ``driving at night''). For \methodconcept{}, we report results using either the optimized concept (``$\mathsf{S}^*$'') alone or combined with a target style description (\eg, ``$\mathsf{S}^*$ at night'').}
    \label{tab:prompt_const}
    \label{tab:prompt_soft}
\end{table}

\begin{figure}
\begin{subfigure}{0.495\linewidth}
\includegraphics[width=\linewidth]{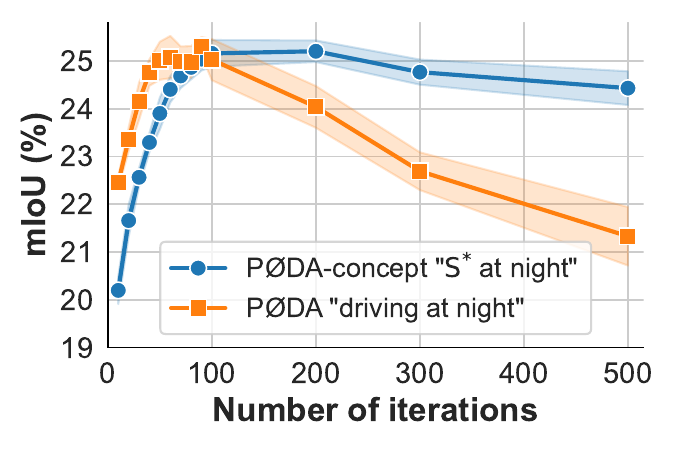}
\caption{ACDC Night}
\label{fig:sub1}
\end{subfigure}\hfill%
\begin{subfigure}{0.495\linewidth}
\includegraphics[width=\linewidth]{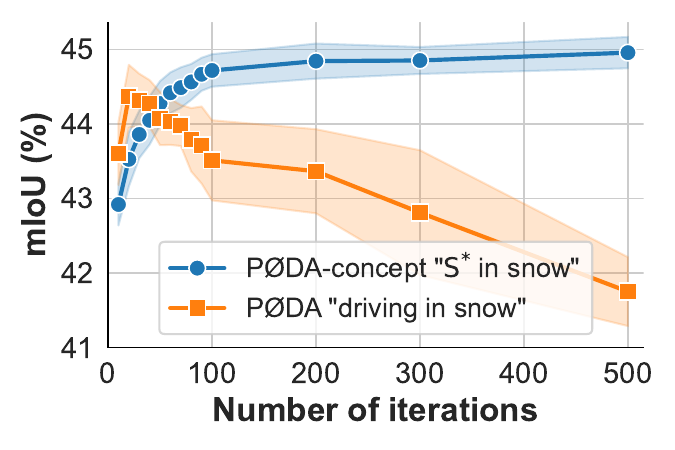}
\caption{ACDC Snow}
\label{fig:sub2}
\end{subfigure}
\medskip
\begin{subfigure}{0.495\linewidth}
\includegraphics[width=\linewidth]{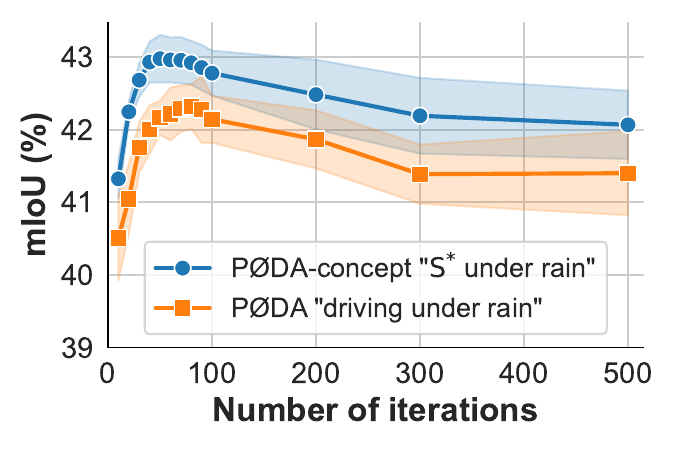}
\caption{ACDC Rain}
\label{fig:sub3}
\end{subfigure}\hfill%
\begin{subfigure}{0.495\linewidth}
\includegraphics[width=\linewidth]{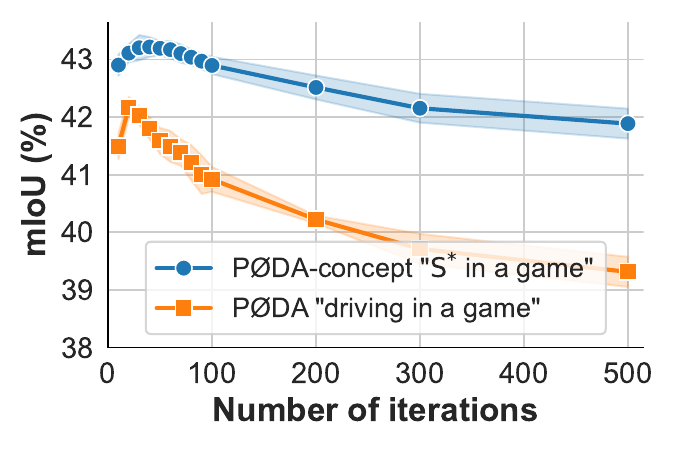}
\caption{GTA5}
\label{fig:sub4}
\end{subfigure}
\caption{
\small\textbf{Effect of the number of optimization iterations.} mIoU (\%) performance of \method{} and \methodconcept{} as a function of the number of PIN optimization iterations. For example, 200 iterations means that PIN is optimized for 200 steps for both \method{} and \methodconcept{}; the resulting styles are saved and then used to fine-tune the segmentation model in each case. Results are shown for zero-shot adaptation from Cityscapes to ACDC subsets \textemdash{} (a)~night, (b)~snow, (c)~rain \textemdash{} as well as to (d)~GTA5. Reported values are averaged over 5 runs.
}
\label{fig:iter_soft_vs_hard}
\end{figure}

\smallskip\noindent\textbf{Concept optimization from source data.~}
As introduced in~\cref{sec:concept_learning}, since the images of the source and unseen target domains share similar general content, we propose to optimize the latter using the source images instead of relying on text. This is done by optimizing a word embedding such that the text latent embedding is close to the source images' embeddings in the bimodal space~(See~\cref{eqn:concept_learning}). 
Results of \methodconcept in~\cref{tab:prompt_soft} show superior performance compared to all variants of \method.
Using a concept optimized from images better encodes the semantic content common to the data and results in better mined styles.
\noindent Interestingly, using $\concept{}$ alone improves the performance w.r.t. source-only. This case is similar to the use of \method with ``driving'', yet $\concept{}$ is more data-specialized. Note that for severe style shift, \eg, night, the mIoU still significantly lags behind $\concept{}$ + ``at night''.

~\cref{fig:iter_soft_vs_hard} shows that including an optimized concept in the prompt not only improves performance but also makes style mining less sensitive to the number of iterations. Indeed, optimizing towards a prompt including $\concept{}$ restricts the search space to a region encoding the semantics of the images, which reduces the risk of drifting from the content. For a high number of iterations, the performance is significantly harmed when using a concept expressed through language (\ie, ``driving'').
Interestingly, using an optimized $\concept{}$, the same observations hold, however the decrease is significantly smaller. We conjecture that in both cases, this might be due to over-stylization~\citep{kwon2022clipstyler}.

In short, using an optimized concept along with the style condition has two advantages: better performance and lower sensitivity to the optimization iterations.

\subsection{\methodone}
\label{sec:one_shot_pin}

We now evaluate our one-shot adaptation, \methodone, which instead of a target prompt leverages a target image. \methodone{} relies on the same mining strategy (\cref{sec:stylemining}) but using a target embedding derived from a target image $\mb{I}_\text{t}$ (cf.~\cref{sec:photodrivennorm}). As in prior experiments, we always report the average performance over 5 runs though here, using a different $\mb{I}_\text{t}$ randomly sampled from the target training set.
Because one-shot adaptation enables nuanced settings that are hard to verbalize, we explore varying levels of granularity in the following experiments.

\smallskip\noindent{}\textbf{Discrete domains.~} \cref{tab:OSUDA_PODA} reports performance on the same discrete domains (\ie, rain, night, etc.) as done previously. Since \methodone{} has access to a target image, we also report a vanilla AdaIN-adaptation baseline which seeks direct alignment with the style of the target image. 
As expected, \methodone{} significantly outperforms AdaIN, a result we attribute to our mining strategy providing greater diversity. 
Conversely, \methodone{} is outperformed almost consistently by its \mbox{0-shot}~counterpart, \method{}. 
We conjecture that this 
is due to the prompts of \method{} better capturing the overall appearance of the target domain, 
whereas randomly selected target images may not adequately represent the entire distribution of appearances in the target domain.
This interpretation is supported by the high variance of \methodone{}, suggesting that mined styles are highly dependent on the choice of the target image.

\begin{table}[t]
	\setlength{\tabcolsep}{0.01\linewidth}
	\def\arraystretch{1.3}
	\centering
	\resizebox{1.0\linewidth}{!}{
		\newcommand{\MPrompt}[1]{\cellcolor{gray!34}``#1''}
		\begin{tabular}{cccccc}
			\toprule
    		&Method & ACDC Night & ACDC Snow & ACDC Rain & GTA5\\
    		\midrule 
			&source-only & 18.31 & 39.28 & 38.20 & 39.59 \\
			\midrule
			\multirow{2}{*}{\rotatebox{90}{\textbf{0-shot}\hspace{-0.1em}}}&\multirow{2}{*}{\method{}} & \small{\MPrompt{driving at night}} & \small{\MPrompt{driving in snow}} & \small{\MPrompt{driving under rain}} & \small{\MPrompt{driving in a game}}\\
			& & {25.03}\vartn{0.48} & {43.90}\vartn{0.53} & 42.31\vartn{0.55} & {41.07}\vartn{0.48}\\
			\midrule
			\multirow{2}{*}{\rotatebox{90}{\textbf{1-shot}\hspace{-0.1em}}} & AdaIN & 18.15\vartn{3.65} & 40.25\vartn{1.72} & 42.61\vartn{0.80}  & 40.56\vartn{0.97} \\
			& \methodone{} & \textbf{24.29}\vartn{1.07}  & \textbf{43.66}\vartn{0.99} & \textbf{42.66}\vartn{1.21} & \textbf{41.02}\vartn{1.86} \\

			\bottomrule
		\end{tabular}
	}
	\smallskip\caption{
    \small\textbf{\methodone{} on discrete domains.} mIoU (\%) performance for 0-shot adaptation using a prompt (\method) and two 1-shot variants \textemdash AdaIN and \methodone \textemdash that leverage a single target image. 
    }
	\label{tab:OSUDA_PODA}
\end{table}

\smallskip\noindent{}\textbf{Continuous domains.~}To better show the interest of \methodone{} for nuanced target conditions, harder to verbalize, we report performance on timelapse data. As such data are lacking in the driving domain, we use CoMoGAN~\citep{pizzati2021comogan}, a continuous image-to-image translation method trained on Waymo Open dataset~\citep{sun2020scalability}, where the translation is parametrized by \mbox{$\phi \in [0,\pi]$}, mapping the sun elevation in \mbox{$[+30^{\circ},-40^{\circ}]$} where negative values indicates that the sun is below the horizon~(\ie,~nightime).
Using CoMoGAN, we translate Cityscapes training and validation sets for different $\phi$ values.
For each $\phi$, we leverage \methodone{} using for adaptation a random image from the $\phi$-translated training set.

In~\cref{fig:comogan}, we plot the Cityscapes validation performance for each $\phi$ value, which shows that \methodone{} consistently outperforms the source-only model. 
In the latter plot, we also report performance of the \mbox{0-shot} \method using various prompts constructed with \texttt{``driving at <time>''}. 
Interestingly, prompts referring to daytime (\eg, ``driving at noon'' or ``at 2pm'') perform better than other prompts when the sun is higher (\ie, $\phi < \pi/2$) while night prompts (\eg, ``driving at 10pm'' or ``at midnight'') are better after sunset (\ie, $\phi > \pi/2$).
Still, for all $\phi$ values, \methodone{} consistently outperforms \method{} regardless of the prompt used. Such results hint that using an image as adaptation guidance can lead to better performance for peculiar conditions which are harder to describe with text.

\begin{figure}[t]
	\centering
	\includegraphics[width=1.0\linewidth]{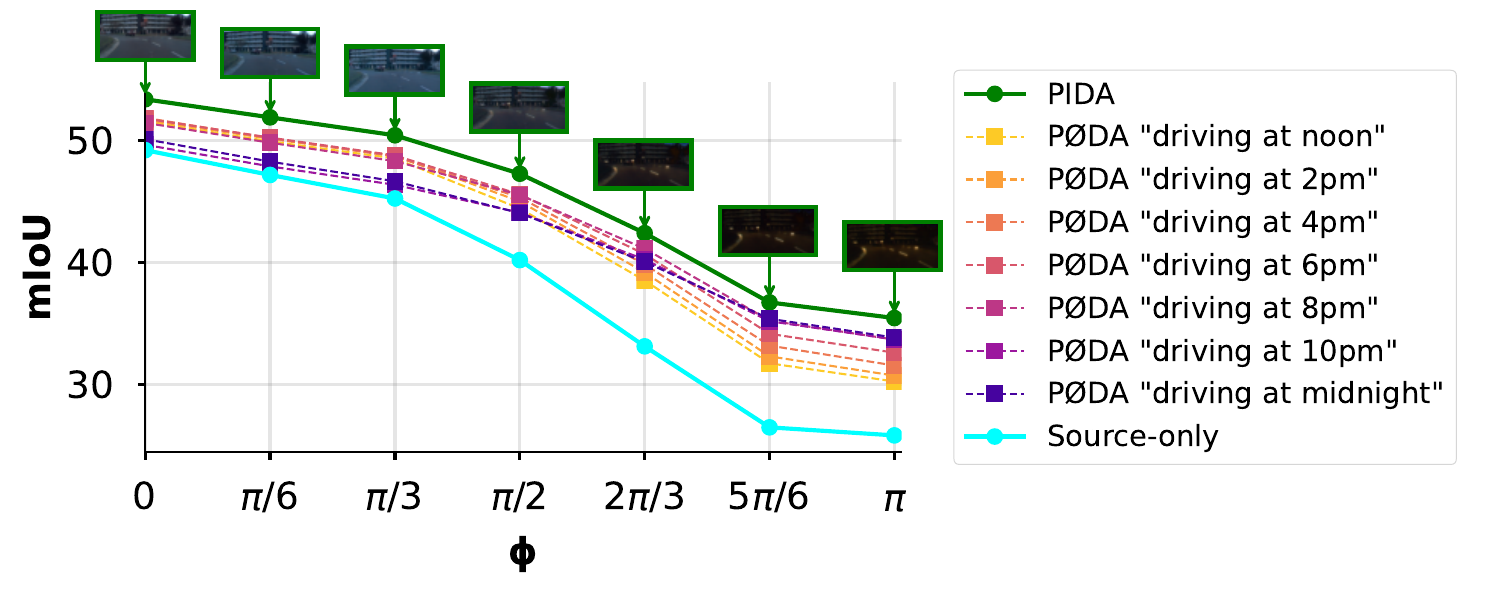}
	\caption{
    \small\textbf{\methodone{} on continuous time-of-day domains.} Average performance (mIoU \%) over 5 runs on $\phi$-translated Cityscapes for source-only, \methodone, and \method{} with different prompts. Example one-shot target images used for \methodone are shown as insets.
    }
	\label{fig:comogan}
\end{figure}

\section{Discussion}
\label{sec:discussion}

\noindent\textbf{Generalization with \method{}.~}
Inspired by the observation that some unrelated prompts improve performance on target domains (see \cref{tab:text_effect_RN50_irrelevant}), we study how \method{} can benefit from general style augmentation.
	First, we coin \makebox{``Source-only-G''} the generalized source-only model where 
	we augment features by shifting the per-channel $(\boldsymbol{\mu},\boldsymbol{\sigma})$ with Gaussian noises sampled for each batch of features, such that the signal to noise ratio is 20 dB. 
	{This source-only variant takes inspiration {from~\citep{fan2023towards}} where simple perturbations of feature channel statistics could help achieve SOTA generalization performance in object detection.}
	\cref{tab:generalization} shows that {Source-only-G} always improves over Source-only, demonstrating a generalization capability.
	When applying our zero-shot adaptation on Source-only-G (denoted~\makebox{``\method{}-G''}), target performance again improves -- always performing best on the desired target. 

 To further boost the performance, we perform style mixing.~\citet{wu2022style} showed that for OSUDA, mixing the source feature statistics and those of the single target image brings generalization effects. Later, ~\citet{fahes2024simple} showed that there is no need for the target image: mixing original and PIN-augmented statistics helps generalization. We follow the same strategy here.
The mixed statistics $\bs{\mu}_{\text{mix}},\bs{\sigma}_{\text{mix}}$ are given by:
\begin{align}
	\bs{\mu}_{\text{mix}} = \bs{\alpha}\odot\trg{\bs{\mu}} + (1-\bs{\alpha})\odot\src{\bs{\mu}}\,, \\
	\bs{\sigma}_{\text{mix}} = \bs{\alpha}\odot\trg{\bs{\sigma}} + (1-\bs{\alpha})\odot\src{\bs{\sigma}}\,, 
\end{align}
where $\bs{\alpha} \in \mathbb{R}^{c}$ are per-channel mixing weights uniformly sampled in $[0,1]$; $\odot$ denotes element-wise multiplication. Addition and substraction are also element-wise. Finally, the augmented features are computed as follows:
\begin{align}
	\mb{f}_{\text{s}\shortrightarrow\text{t}} = \texttt{AdaIN}(\src{\mb{f}}, \bs{\mu}_{\text{mix}}, \bs{\sigma}_{\text{mix}}), 
\end{align}
with $\texttt{AdaIN}$ defined in \cref{eqn:adain}.
  
\begin{table}[t]
    		\setlength{\tabcolsep}{0.01\linewidth}
    \centering
	\resizebox{1.0\linewidth}{!}{
  \begin{tabular}{lllll}
    \toprule
    Method & Night & Snow & Rain & GTA5\\
    \midrule
    Source-only & 18.31 & 39.28 & 38.20 & 39.59\\ 
    Source-only-G & 21.07 & 42.84 & 42.38 & 41.54\\
    \method{}-G & \textbf{24.86}\vartn{0.70} & 44.34\vartn{0.36} & 43.17\vartn{0.63} & 41.73\vartn{0.39}\\
    \method{}-G+style-mix & 24.18\vartn{0.23} & \textbf{44.46}\vartn{0.34} & \textbf{43.56}\vartn{0.46} & \textbf{42.98}\vartn{0.12} \\
    \bottomrule
  \end{tabular}
  }
  \smallskip\caption{
  \small\textbf{Generalization with \method{}.} The ``Source-only-G'' model applies a domain generalization technique by perturbing low-level feature statistics with Gaussian noise. Building on this, \method{}-G (\method{} applied to Source-only-G) yields further performance gains. \makebox{`style-mix'} refers to a style mixing strategy inspired by \citep{wu2022style}.
  }
  \label{tab:generalization}
\end{table}

\smallskip\noindent\textbf{\method{} with different architectures.~} {We show in \cref{tab:fpn} consistent gains brought by \method{} using other backbone (RN101~\citep{he2016deep}) and segmentation head (semantic FPN~\citep{kirillov2019panoptic}).}

\begin{table}[!t]
	\setlength{\tabcolsep}{0.01\linewidth}
	\centering
	\resizebox{1.0\linewidth}{!}{
		\begin{tabular}{llllll}
			\toprule
			Backbone & Method & Night & Snow & Rain & GTA5\\
			\midrule
			\multirow{2}{*}{Sem.\,FPN} & src-only &  18.10 & 35.75 & 36.07 & 40.67 \\
			& \method& \textbf{21.48}\vartn{0.15} & \textbf{39.55}\vartn{0.13} & \textbf{38.34}\vartn{0.29} & \textbf{41.59}\vartn{0.24} \\
			\midrule
			\multirow{2}{*}{DLv3+} & src-only & 22.17 & 44.53 & 42.53 & 40.49 \\
			& \method & \textbf{26.54}\vartn{0.12} & \textbf{46.71}\vartn{0.43} & \textbf{46.36}\vartn{0.20} & \textbf{43.17}\vartn{0.13} \\
			\bottomrule 
		\end{tabular}
	}
	\smallskip\caption{
    \small\textbf{\method{} with different architectures.} We evaluate \method using two segmentation architectures: Semantic FPN (`Sem.\,FPN') with a ResNet-50 backbone and DeepLabV3+ (`DLv3+') with a ResNet-101 backbone. This comparison highlights the effectiveness of \method across different model designs.
    }
	\label{tab:fpn}
\end{table}

\smallskip\noindent\textbf{Topic importance.~} In the era of large multimodal foundation models achieving unprecedented performance on similar tasks~\citep{bai2025qwen2,kirillov2023segment}, we believe our work tackles a complementary and still critical problem space.
Specifically, we argue that making domain adaptation faster and more accessible for smaller models---such as ResNet-50 with DeepLabV3+, which are widely used in practice due to their efficiency and deployability---is of significant importance.
Moreover, lightweight adaptation is highly practical in cases where large-scale retraining is not viable and fast inference is required, such as in resource-constrained environments or real-time applications.
Lastly, while large-scale models are less prone to distribution shifts, we argue that understanding domain gaps in the context of smaller models could reveal a deeper theoretical understanding of domain adaptation mechanisms more broadly.

\smallskip\noindent\textbf{Limitations.~}
PIN assumes that domain shift is entirely encoded in the visual style, therefore overlooking any camera and content variations. Another underlying assumption is that the style is entirely modeled by the channel-wise mean and standard deviations.
While both assumptions are widely adopted for image translation and domain adaptation, they further limit the applicability of PIN. Investigating this is essential for deeper understanding of PIN’s role and its synergy with other adaptation methods.

\smallskip\noindent\textbf{Future directions.~}
Future research could explore integrating our framework with online techniques like test-time adaptation (TTA)~\citep{wang2021tent}, which enables on-the-fly learning during inference. We envision a dual-stage adaptation pipeline where our ``offline'' approach would provide a robust foundation by adapting to broad target characteristics through generic prompts (e.g., \textit{driving at night}), while TTA would handle instance-specific shifts. 
Combining of the latter would allow to benefit from both general domain alignment and fine-grained, real-time adjustments. Investigating this synergy is a promising direction for achieving more resilient generalization in dynamic environments.

\section{Conclusion}
{In this work, we leverage the CLIP model to make possible a new challenging task of domain adaptation using a single VL embedding.
We propose a cost-effective feature augmentation mechanism that adjusts the style-specific statistics of source features to synthesize augmented features in the target domain, guided by a single VL embedding coming either from a prompt in natural language, a partially optimized prompt, or a single target image.
Extensive experiments, particularly for semantic segmentation, have proven the effectiveness of our framework for semantic segmentation in particular. They also show its applicability to other tasks and various backbones.
Our extensive experiments on the Cityscapes and ACDC datasets—which reflect real-world driving conditions, including challenging scenarios such as night, snow, and rain—demonstrate the practical relevance of our method in safety-critical applications. By relying only on a language prompt or a single image, our approach significantly reduces the dependency on large-scale target data collection, representing a step toward scalable and efficient deployment in real-world autonomous driving systems.
Our line of research aligns with the collective efforts of the community to leverage large-scale pre-trained models (so-called ``foundation models''~\citep{bommasani2021opportunities}) for 
data- and label-efficient training of perception models for real-world applications. Future research may explore hybrid solutions for robust segmentation based on both large systems and our lightweight adaptation strategy.}

\smallskip\noindent\textbf{Data availability statement.} All the datasets used in this paper are publicly available.

\smallskip\noindent
\textbf{Acknowledgment.~} This work was partially funded by French project SIGHT (ANR-20-CE23-0016) and the European Union under grant agreement No.~101070617.

\appendix

\section{\method{} on other tasks}
\label{sec:extensiontasks}
{\method{} operates at the features level, which makes it task-agnostic. We show in the following the effectiveness of our method for object detection and image classification.}

\smallskip\noindent\textbf{\method{} for Object Detection.~}
We report results in~\cref{tab:poda_od_results} from straightforwardly applying~\method{} to object detection.
Our Faster-RCNN~\citep{ren2015faster} models, initialized with two CLIP-pretrained backbones, are trained on two source datasets, either Cityscapes or the Day-Clear split in Diverse Weather Dataset (DWD)~\citep{wu2022single}.
We report adaptation results on Cityscapes-Foggy~\citep{sakaridis2018semantic} and four other conditions in DWD.
For zero-shot feature augmentation in~\method{}, we use simple prompts and take the default optimization parameters in previous experiments.
~\method{} obtains on par or better results than
UDA methods~\citep{chen2021scale,rezaeianaran2021seeking} {(which use target images)} and domain generalization methods~\citep{fan2023towards,wu2022single,vidit2023clip}. We also experimented with YOLOF~\citep{chen2021you} for object detection in CS$\shortrightarrow$Foggy;~\method{} reaches $35.4\%$, improving $1.5\%$ from the source-only model.
These  
results open up potential combinations of~\method{} with generalization techniques like~\citep{fan2023towards} and~\citep{vidit2023clip} for object detection.

In the Cityscapes-Foggy experiment, we freeze \textit{Layer1} of the backbone, while in the DWD experiments we additionally freeze \textit{Layer2}. The remaining layers are fine-tuned as commonly done. The code for~\method{} in object detection was built upon the MMDetection library.\footnote{\url{https://github.com/open-mmlab/mmdetection}}

\begin{table}[t!]
    \setlength{\tabcolsep}{0.01\linewidth}
    \centering
    \resizebox{1.0\linewidth}{!}{
    \begin{tabular}{lcccccc}
        \toprule
        &&\multirow{3}{*}{\makecell{CS$\shortrightarrow$ CS\\Foggy}} & \multicolumn{4}{c}{DWD-Day Clear $\shortrightarrow$} \\ \cmidrule(lr){4-7}
        Method & Target &  &\makecell{Night\\Clear} & \makecell{Dusk\\Rainy} & \parbox{0.8cm}{Night\\Rainy}&\parbox{0.8cm}{Day\\Foggy}  \\
         \midrule
         DA-Faster~\citep{chen2021scale} & \checkmark & 32.0&-&-&-&-\\
         ViSGA~\citep{rezaeianaran2021seeking} & \checkmark & 43.3&-&-&-&-\\
         NP+~\citep{fan2023towards} & \xmark & 46.3&-&-&-&-\\
         S-DGOD~\citep{wu2022single} & \xmark & -&36.6&28.2&16.6&33.5\\
         CLIP The Gap~\citep{vidit2023clip} & \xmark & -&36.9&32.3&18.7&38.5\\
         \method{} & \xmark & \textbf{47.3}& \textbf{43.3} & \textbf{40.2} & \textbf{20.5} & \textbf{43.1}
         \\
         \bottomrule
    \end{tabular}
    }
    \smallskip\caption{\small \textbf{\method for object detection (mAP\%)}. For Cityscapes$\shortrightarrow$Cityscapes-Foggy adaptation, the backbone is ResNet-50, while it is ResNet-101 for adaption from DWD-Day-Clear to other conditions in DWD.
    }
    \label{tab:poda_od_results}
\end{table}

\smallskip\noindent \textbf{\method{} for Image Classification.}
We show that \method{} can be also applied for image classification. We use the same augmentation strategy to adapt a linear probe on top of CLIP-RN50 features.
In a first experiment we train a linear classifier on the features of CUB-200 dataset~\citep{wah2011caltech} of 200 real bird species. We then perform zero-shot adaptation to classify bird paintings of CUB-200-Paintings dataset~\citep{wang2020progressive} using the single prompt \makebox{``Painting of a bird''}. \method{} improves the source-only performance from 28.90 to 30.91\vartn{0.69}. In our second experiment, we address the color bias in Colored MNIST~\citep{arjovsky2019invariant}. While for training, \textcolor{red}{even} and \textcolor{blue}{odd} digits are colored \textcolor{red}{red} and \textcolor{blue}{blue} respectively, the test digits are randomly colored. We augment training digit features using the \makebox{``Blue digit''} and \makebox{``Red digit''} prompts for \textcolor{red}{even} and \textcolor{blue}{odd} digits respectively, and create a separate set for each one to prevent styles from leaking, \ie, to avoid trivially using ``red'' styles coming from \textcolor{red}{even} digits to augment \textcolor{blue}{odd} digits features and vice versa. 
Again, \method{} brings significant improvement over the source-only model (64.16\vartn{0.41} \vs 55.83).

\section{Additional experiments}
\label{sec:additional_exp}

\noindent\textbf{Effect of style mining initialization.}
In our feature optimization step, we initialize $(\bs{\mu},\bs{\sigma})$ with $(\mu(\src{\mb{f}}),\sigma(\src{\mb{f}}))$.
In~\cref{tab:init_effect}, we report results using different initialization strategies. 
Starting from pre-defined or random initialization, instead of from original statistics, degrades badly the performance.
As we do not use any regularization term in the CLIP cosine distance loss, we argue that initializing the optimized statistics with those of the source images is a form of regularization, favoring augmented features in a neighborhood of $\src{\emb{\mb{f}}}$ and better preserving the semantics.

\begin{table}[t]
	\centering
	\resizebox{0.7\linewidth}{!}{
        \begin{tabular}{ccr}
		\toprule
		$\bs{\mu}^{0}$ & $\bs{\sigma}^{0}$ & mIoU\quad\quad \\
		\midrule
		$\mu(\src{\mb{f}})$ & $\sigma(\src{\mb{f}})$ & \textbf{25.03}\vartn{0.48} \\
        $\mathbf{0}$ & $\mathbf{1}$ & 8.59\vartn{0.82} \\
	    $\sim \mathcal{N}(\mathbf{0},\mathbf{I})$ & $\sim \mathcal{N}(\mathbf{0},\mathbf{I})$  & 6.80\vartn{0.92} \\
		\bottomrule
	\end{tabular}}
	\smallskip\caption{\small \textbf{Effect of style initialization.} Performance (in mIoU) of \method{} on ACDC-Night val set (Cityscapes as source), with different initializations of style statistics. Starting from source images' statistics works substantially better.}
\label{tab:init_effect}
\end{table}

\smallskip\noindent\textbf{Diversity of optimized statistics.}
To verify that the statistics --- optimized for the same number of iterations with the same $\prompt$ but from different starting anchor points $\emb{\src{\mb{f}}}$ --- are diverse, we show in~\cref{fig:learned_parameters_boxplots} the boxplots of optimized parameters on the first $20$ channels of $\stot{\mb{f}}$ (for prompt ``driving at night'').

 \begin{figure}[t!]
     \centering
         \includegraphics[width=\linewidth]{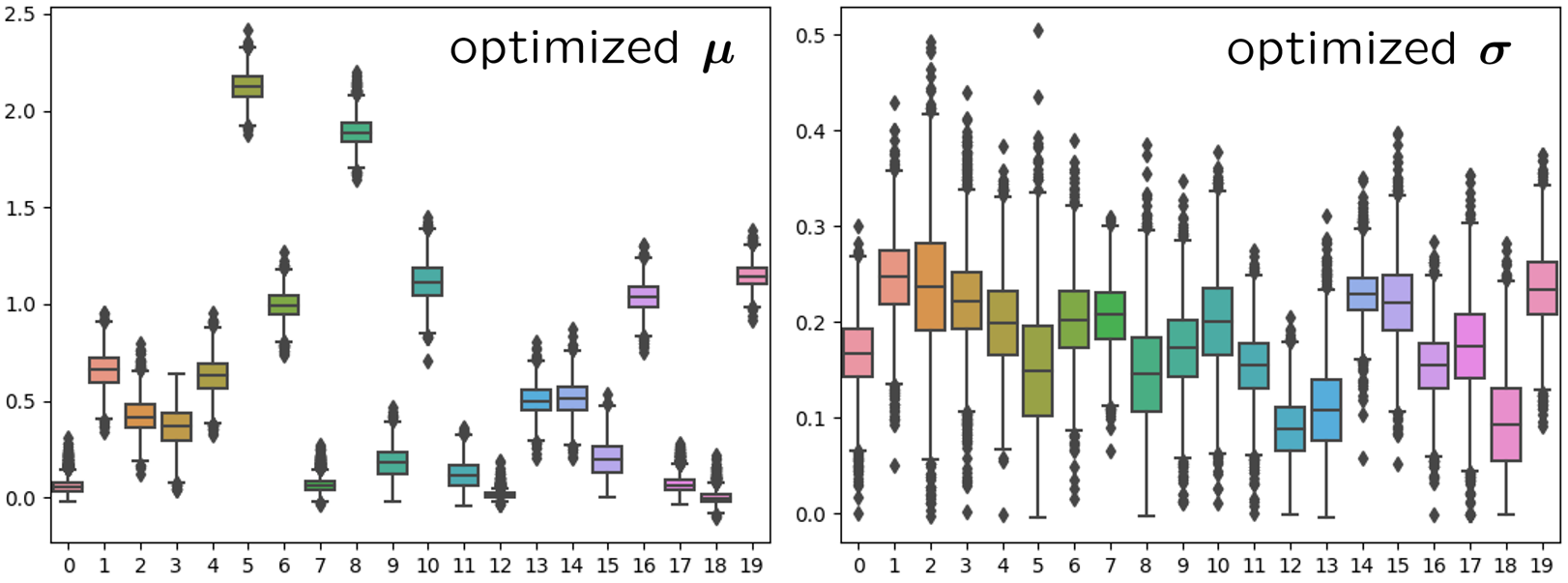}
     \caption{\small \textbf{Per-channel optimized statistics.} Distributions of the first $20$ channels of the optimized statistics of $\bs{\mu}$ (\textit{Left}) and $\bs{\sigma}$ (\textit{Right}). Each boxplot shows the interquartile range (IQR) that contains 50\% of the data: Its bottom and top edges delimit the first and third quartiles respectively. The horizontal line inside the box denotes the data's median. The whiskers extend from the edges of the box to the furthest point within $1.5$ times the IQR, in each direction. Outlier points beyond these limits are individually plotted (diamonds).}
    \label{fig:learned_parameters_boxplots}
\end{figure}

\smallskip\noindent\textbf{Training from scratch on augmented features.} In \method, we start with a source-only trained model (\cref{algo:PODA_overall}, line 1) then we fine-tune it on augmented features (\cref{algo:PODA_overall}, line 4). This is the general setting for domain adaptation. However, since our method performs domain adaptation under the assumption of label preservation, we also experimented training the model from scratch on augmented features. The results (\cref{tab:from_scratch}) show the importance of the first, source-only training step.

\begin{table}[t]
	\setlength{\tabcolsep}{0.015\linewidth}
	\centering
	\begin{tabular}{lcccc}
		\toprule
		Method & Night & Snow & Rain & GTA5\\
		\midrule
        \method no src. pretrain & 22.46 & 36.73 & 39.70 & 39.57\\
        \midrule
        \method& \textbf{25.03} & \textbf{43.90} & \textbf{42.31} & \textbf{41.07}\\	    
        \bottomrule
	\end{tabular}
	\smallskip\caption{\small \textbf{Importance of source-only pre-training}. Semantic segmentation performance (mIoU \%) of \method \vs its variant without source-only training, when adapting from Cityscapes to ACDC Nigt/Snow/Rain and to GTA5.}
\label{tab:from_scratch}
\end{table}

\bibliography{sn-bibliography}

\end{document}